\documentclass[a4paper,fleqn]{cas-dc}
\usepackage[authoryear]{natbib}
\usepackage{multirow}
\usepackage{subfig}
\usepackage{wrapfig}
\usepackage{balance}
\usepackage{threeparttable}
\def\tsc#1{\csdef{#1}{\textsc{\lowercase{#1}}\xspace}}
\tsc{WGM}
\tsc{QE}

\begin{document}
\let\WriteBookmarks\relax
\def\floatpagepagefraction{1}
\def\textpagefraction{.001}
\let\printorcid\relax

\shorttitle{A dual-stream recurrence-attention network with global--local awareness}

\shortauthors{Li et al.}  

\title[mode = title]{A dual-stream recurrence-attention network with global--local awareness for emotion recognition in textual dialog}

\tnotetext[1]{Accepted by Engineering Applications of Artificial Intelligence (EAAI). DOI: 10.1016/j.engappai.2023.107530. Received 3 August 2023; Received in revised form 13 November 2023; Accepted 14 November 2023. \textit{E-mail address:} lijfrank@hust.edu.cn (J. Li), wangxiaoping@hust.edu.cn (X. Wang), zgzeng@hust.edu.cn (Z. Zeng).}

\author[1,2,3,4]{Jiang Li}
\cormark[1]
\author[1,3,4]{Xiaoping Wang}
\cormark[1]
\cortext[1]{Corresponding author at: School of Artificial Intelligence and Automation, Huazhong University of Science and Technology (HUST), Wuhan 430074, China.}
\author[1,3,4]{Zhigang Zeng}
\address[1]{School of Artificial Intelligence and Automation, Huazhong University of Science and Technology (HUST), Wuhan 430074, China}
\address[2]{Institute of Artificial Intelligence, Huazhong University of Science and Technology (HUST), Wuhan 430074, China}
\address[3]{Key Laboratory of Image Processing and Intelligent Control, Ministry of Education, Wuhan 430074, China\\}
\address[4]{Hubei Key Laboratory of Brain-inspired Intelligent Systems, Wuhan 430074, China}

\begin{abstract}
In real-world dialog systems, the ability to understand the user's emotions and interact anthropomorphically is of great significance. Emotion Recognition in Conversation (ERC) is one of the key ways to accomplish this goal and has attracted growing attention. How to model the context in a conversation is a central aspect and a major challenge of ERC tasks. Most existing approaches struggle to adequately incorporate both global and local contextual information, and their network structures are overly sophisticated. For this reason, we propose a simple and effective Dual-stream Recurrence-Attention Network (DualRAN), which is based on Recurrent Neural Network (RNN) and Multi-head ATtention network (MAT). DualRAN eschews the complex components of current methods and focuses on combining recurrence-based methods with attention-based ones. DualRAN is a dual-stream structure mainly consisting of local- and global-aware modules, modeling a conversation simultaneously from distinct perspectives. In addition, we develop two single-stream network variants for DualRAN, i.e., SingleRANv1 and SingleRANv2. According to the experimental findings, DualRAN boosts the weighted F1 scores by 1.43\% and 0.64\% on the IEMOCAP and MELD datasets, respectively, in comparison to the strongest baseline. On two other datasets (i.e., EmoryNLP and DailyDialog), our method also attains competitive results.
\end{abstract}
\maketitle

\section{Introduction}\label{introduction}
Emotion recognition is a promising application and has received a great deal of attention from academics in recent years. Emotion Recognition in Conversation (ERC) is a subfield of emotion recognition with special scenarios. ERC offers plenty of potential application scenarios. Examples include (1) in disease diagnosis, to assist the doctor in diagnosing a disease by identifying the emotional status of the patient when talking with a psychologist; (2) in opinion mining, to enhance the service quality of the governmental department or organization by analyzing the public's emotional experience towards the policy or service; (3) in dialog generation, to raise the usability of the system by injecting emotions into the given model; and (4) in recommender systems, to infer the potential preferences of a target user by recognizing the user's emotional states while chatting with customer service.

\begin{figure}[htbp]
    \centering
    \includegraphics[width=3.0in]{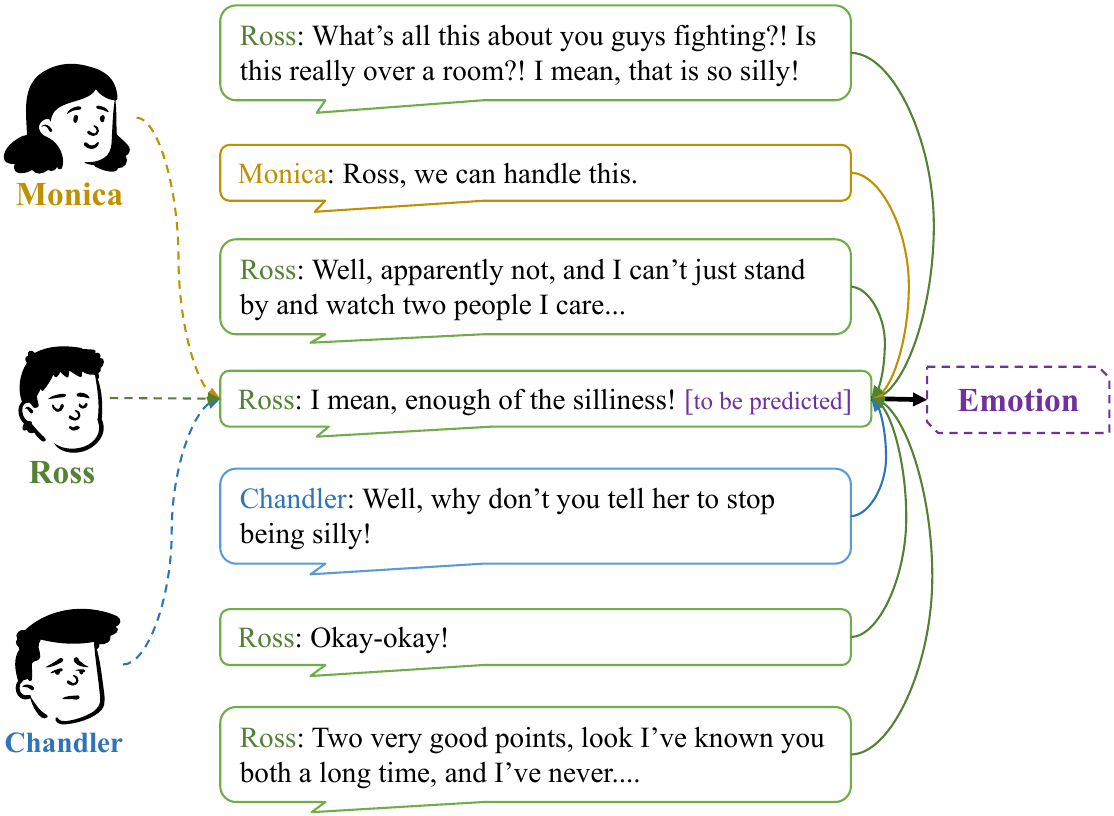}
    \caption{An example of emotion recognition in conversation. The emotion of the utterance to be predicted is influenced by the utterance itself, the contexts (solid lines), and the identities of speakers (dashed lines).}
    \label{fig:example}
\end{figure}

Distinct from general emotion recognition, ERC not only focuses on the utterance itself but also demands that the contexts of the utterance is sufficiently understood~\cite[]{song2023sunet}. Figure~\ref{fig:example} is a general flow that embodies the ERC task. The input of this task is a sequence of utterances in the conversation and the corresponding speakers, and its output is the emotions of these utterances. The emotion of the utterance to be predicted is affected by the utterance itself, the contexts, and the identities of speakers. With the rapid deployment and development of human-computer interaction, there is an urgent need to engage machines that can interact more naturally and humanely with humans. As a result, the importance of building conversational systems that can understand human emotion and intention has grown significantly~\cite[]{peng2020humanmachine}. The development of ERC, which fits the above-mentioned usage scenarios for dialog systems, is urgent and has attracted increasing research in natural language understanding communities.

Plenty of efforts have been made in context-based modeling, and these ERC models fall into three main categories: recurrence-based approaches, Transformer-based approaches, and graph-based approaches. Recurrence-based methods are sensitive to the order of utterances and treat these utterances as a temporal sequence. COSMIC~\cite[]{ghosal2020cosmic} was a conversational emotion recognition framework based on commonsense knowledge guidance, claiming to alleviate the problems of emotion shift and similar emotion. AGHMN~\cite[]{jiao2020real} was a conversational emotion recognition model based on Gated Recurrent Unit (GRU)~\cite[]{chung2014empirical} for building a memory bank to capture historical contexts and summarize memories to extract critical information. DialogueCRN~\cite[]{hu2021dialoguecrn} enhanced the extraction and integration of emotional cues and was a contextual reasoning network based on cognitive theory. CauAIN~\cite[]{zhao2022cauain} introduced commonsense knowledge as a cue for emotion cause detection in conversation, explicitly modeling intra- and inter-speaker dependencies. But these models are struggling to capture the global contextual information of the utterance. To solve this problem, some Transformer-based and graph-based methods have been proposed successively. Benefiting from the advantage of multi-head attention, Transformer-based methods allow for the consideration of long-range contextual information. HiTrans~\cite[]{li2020hitrans} was a context- and speaker-aware model based on the hierarchical Transformer~\cite[]{vaswani2017attention}. DialogXL~\cite[]{shen2021dialogxl} was a pioneering work based on the pre-trained language network XLNet~\cite[]{yang2019xlnet}, which modified the network structure of XLNet to better model conversational emotion data. CoG-BART~\cite[]{li2022contrast} was a conversational emotion recognition model that applied the encoder-decoder model BART~\cite[]{lewis2020bart} as a backbone network. Graph-based methods, similar to Transformer-based methods, are capable of modeling contexts of the utterance from a global perspective. SKAIG-ERC~\cite[]{li2021pastpresent} utilized a psychological-knowledge-aware interaction graph to model the historical context and commonsense knowledge of utterance. I-GCN~\cite[]{nie2022igcn} first represented conversations at different times using a graph structure and then simulated dynamic conversational processes using an incremental graph structure to capture both semantic correlation information of utterances and time-varying information of conversations.

However, these methods either focus on the local sequence information of utterance or the global association information of utterance, ignoring the combination of local and global information. Although recurrence-based ERC methods can extract the temporal sequence information of dialog sequences, they tend to capture the nearest contextual information (i.e., focusing on the extraction of local information) and have difficulty in capturing long-range contextual information. Transformer-based and graph-based ERC methods can alleviate these problems, but they do not take into account the temporal information of utterance and have difficulty in adequately capturing the local information of utterance. In addition, some ERC models have an overly complex network structure, such as incorporating commonsense knowledge~\cite[]{ghosal2020cosmic,zhao2022cauain,li2021pastpresent}, including multiple complex modules~\cite[]{jiao2020real,hu2021dialoguecrn,shen2021dialogxl}, adopting an encoder-decoder structure~\cite[]{li2022contrast,zhu2021topic}, etc., consuming numerous computational resources in return for a weak performance gain. 

Therefore, in this paper, we provide a simple and effective dual-stream network structure that explores combining recurrence- and attention-based models so that they complement each other. On the basis of Recurrent Neural Network (RNN) and Multi-head ATtention network (MAT), we construct local-aware and global-aware modules, respectively, and propose a Dual-stream Recurrence-Attention Network (DualRAN) for the ERC task. Furthermore, relying on the local- and global-aware modules in DualRAN, we devise two Single-stream Recurrence-Attention Networks (SingleRAN), which can be regarded as two variants of DualRAN. DualRAN differs significantly from most ERC methods in the following aspects: (1) DualRAN is a dual-stream structure in which two sub-networks can encode information simultaneously; (2) the network structure of DualRAN is simple and directly combines RNN and MAT; and (3) DualRAN can mine both local and global contextual emotional cues, thus more comprehensively extracting contextual information.

The proposed DualRAN is more of a framework as its internal components can be flexibly changed, e.g., employing different types of RNNs, adopting skip connections or not, moving the position of the normalization layer, and even modifying the dual-stream to a single-stream structure. We conduct comparative experiments on four public datasets, and the results show that the proposed structure can lead to competitive performance improvements. With numerous ablation experiments, we explore the validity or impact of different modules on performance, e.g., revealing the effectiveness of local- and global-aware modules, the impact of different RNNs, the effect of speaker identity, the influence of skip connection, and so on. Not only that, we also compare two-stream and one-stream structures, as well as conduct sentiment classification (tri-classification) experiments. Our contribution is as follows:
\begin{itemize}
	\item A simple Dual-stream Recurrence-Attention Network (DualRAN) with global--local-aware capacity is proposed to sufficiently model the contextual dependencies of utterance from both local and global perspectives. DualRAN adopts a dual-stream network structure, consisting mainly of an RNN-based local-aware module and a MAT-based global-aware module. 
	\item To enhance the expressive capacity of RNN, we add two skip connections and a feed-forward network layer to the local-aware module inspired by Transformer architecture. In addition, we change the dual-stream structure in DualRAN to a single-stream one and maintain other components unchanged, providing two single-stream recurrence-attention networks, i.e., SingleRANv1 and SingleRANv2.
	\item We conduct extensive experiments on four widely used benchmark datasets, including comparisons with baselines, comparisons with two SingleRANs, ablation studies for different components, and sentiment classification. The empirical results reveal that the proposed DualRAN can effectively model the ERC dataset and still surpass other models without using external commonsense knowledge.
\end{itemize}

The remaining sections primarily cover related works, methodology, experimental settings, experimental results \& analysis, and conclusion \& prospect. In Section~\ref{related}, we introduce the existing related works. Section~\ref{methodology} corresponds to the methodology of this paper, i.e., we present in detail the DualRAN and its variants proposed in this paper. In Sections~\ref{settings} and ~\ref{results}, we first describe the experimental setup of this work, and then report, discuss, and analyze the experimental results. The last section (i.e., Section~\ref{conclusion}) contains our conclusion and prospect of this work.

\section{Related works}\label{related}
\subsection{Emotion recognition}
Emotion Recognition (ER) is a multidisciplinary research field that covers computer science, cognitive science, psychology, behavior, and sociology. For many years, ER has received broad attention from both academia and industry and has been explored in the domains of computer vision~\cite[]{ullah2022improved,Karnati2023Understanding}, natural language processing~\cite[]{hazarika2020misa}, automatic speech recognition~\cite[]{chen2021anovel}, and signal processing~\cite[]{seal2020eeg}. Emotion recognition data mainly includes two categories: (1) external emotion data, such as facial expression, speech, text; and (2) internal emotion data, such as electroencephalogram, heart rate, and blood pressure. Compared to external emotion data, internal emotion data (typically called physiological signals) requires specialized sensor devices to collect, and some signals, such as electroencephalogram, are challenging to acquire.

In this paper, we focus on the study of context-dependent ER, i.e., Emotion Recognition in Conversation (ERC). Significantly distinct from general context-independent-based ER, the ERC task not only needs to extract the emotion of the current sample but also needs to consider contextual modeling. Limited by the difficulty of dataset collection, current ERC tasks mainly adopt external emotion data as input. In addition, among these data, facial expression and speech usually contain a great deal of noise, while text belonging to artificial data contains cleaner emotional information.

\subsection{Dialog emotion recognition}
Emotion Recognition in Conversation (ERC) is a burning and promising task in recent years. Unlike general emotion recognition, ERC involves conversational context. Emotion Recognition in Conversation (ERC) has attracted extensive research attention owing to its wide range of applications. Depending on the structure of the network, there are mainly recurrence-based methods, Transformer-based methods, and graph-based methods.

Recurrence-Based Methods:
DialogueRNN~\cite[]{majumder2019dialoguernn} was an ERC method based on multiple GRUs that incorporated the speaker information for each utterance to provide more reliable contextual information. COSMIC~\cite[]{ghosal2020cosmic} modeled different aspects of commonsense knowledge by considering mental states, events, actions, and cause-effect relations, and thus extracted complex interactions between personality, events, mental states, intents, and emotions. AGHMN~\cite[]{jiao2020real} mainly consisted of Hierarchical Memory Network (HMN) and Bi-directional GRU (BiGRU), where HMN was used to extract interaction information between historical utterances and BiGRU was used to summarize recent memory and long-term memory with the help of attention weights. DialogueCRN~\cite[]{hu2021dialoguecrn} constructed a multi-turn reasoning module to perform the intuitive retrieving process and conscious reasoning process, thus simulating the cognitive thinking of humans. BiERU~\cite[]{li2022bieru} designed a generalized neural tensor block and a two-channel classifier namely bidirectional emotional recurrent unit to perform contextual feature extraction and sentiment classification. CauAIN~\cite[]{zhao2022cauain} modeled the context of utterance through the perspective of emotion cause detection and was known as a causal aware interaction network. CauAIN consisted of two main cause-aware interactions, i.e., causal cue retrieval and causal utterance retrieval, which were used to find the causal utterance of the emotion expressed by the target utterance. Recurrence-based methods typically treat a conversation as a sequence and can extract temporal order information in the conversation. However, they tend to focus on nearby contextual information and ignore distant one.

Transformer-Based Methods:
HiTrans~\cite[]{li2020hitrans} extracted the contextual information of the utterance with the help of low-level and high-level Transformers, and it extracted the speaker information with the aid of an auxiliary task called pairwise utterance speaker verification. TODKAT~\cite[]{zhu2021topic} was a transformer encoder-decoder structure that combined topic representation and commonsense knowledge for conversational emotion recognition. DialogXL~\cite[]{shen2021dialogxl} was improved in two main ways, the first one was to improve the recurrence mechanism of XLNet from segment-level to utterance-level, and the second one was to replace the original vanilla attention by utilizing dialog-aware self-attention. EmotionFlow~\cite[]{song2022emotionflow} encoded the utterances of speakers by connecting contexts and auxiliary tasks, and it applied conditional random fields to capture sequential features at the emotion level. CoG-BART~\cite[]{li2022contrast} first adopted the utterance-level Transformer to model the long-range contextual dependencies between utterances, then utilized the supervised contrast learning to solve the similar emotion problem, and finally introduced the auxiliary response generation task to enhance the capability of the model to capture contextual information. Despite the fact that Transformer-based methods can model the context from a global perspective, it is challenging to capture the chronological information in a conversation.

Graph-Based Methods: 
KI-Net~\cite[]{xie2021knowledgeinteractive} consisted of two main components to enhance the semantic information of utterances, namely a self-matching module for internal utterance-knowledge interaction and a phrase-level sentiment polarity intensity prediction task. SKAIG-ERC~\cite[]{li2021pastpresent} captured the contextually inferred behavioral action information and future contextually implied intention information leveraging the structure of graph, while the knowledge representation of edges was performed with the help of commonsense knowledge to enhance the emotional expression of the utterance. The approach claimed to model past human actions and future intentions while modeling the mental state of the speaker. S+PAGE~\cite[]{liang2022page} was a graph neural network-based emotion recognition model. The method modeled a conversation as a graph, adding relative location encoding and speaker encoding to the representation of edge weight and edge type, respectively, to better capture speaker- and location-aware conversational structure information. I-GCN~\cite[]{nie2022igcn} first extracted latent correlation information between utterances with an improved multi-head attention module, then focused on mining the correlation between speakers and utterances to provide guidance for utterance feature learning from another perspective. LR-GCN~\cite[]{ren2022lrgcn} first integrated the contextual information and speaker dependencies by utilizing the potential relationship graph network, and then it extracted potential associations between utterances with the multi-head attention mechanism to fully explore the potential relationships between utterances. Analogous to the Transformer-based methods, graph-based methods take into account the global contextual associations and neglect the temporal information due to the structure of the graph.

Although the above approaches model the context to varying extents, the considered perspectives are not comprehensive enough. Recurrence-based ERC focuses on local modeling, making it extremely difficult to consider the context from a global perspective, while Transformer- and graph-based approaches share the problems of often neglecting local and temporal modeling. Additionally, most of the models are overly complex in structure, but the performance gains are not significant enough. Instead, our DualRAN combines the benefits of both recurrence- and Transformer-based methods in a simple way to capture local and global contextual information in the conversation.

\subsection{Machine learning methods}
Our work mainly employs two machine learning methods: Recurrent Neural Network and Multi-Head Attention. Our goal is to combine these two networks in a simplistic way and achieve the best performance. We note that recently there have been new machine learning techniques such as self-distillation~\cite[]{xing2022SelfMatch} and contrastive learning~\cite[]{xiao2023deep}. A number of tasks have achieved unprecedented success using these techniques. In future work, we will consider applying these technologies to our model to improve the performance of ERC.

\subsubsection{Recurrent neural network}
Recurrent Neural Networks (RNNs) are a class of neural network architectures for processing sequential data. The gating mechanism was introduced to solve the problem of gradient explosion or gradient disappearance in the traditional RNN~\cite[]{hochreiter1997long}. \cite{hochreiter1997long} proposed Long and Short Term Memory (LSTM) network to correctly deal with the problem of vanishing gradient. Gated Recurrent Unit (GRU) was proposed by \cite{chung2014empirical} in 2014 and is another classical RNN architecture. RNNs have been widely applied in the field of natural language processing due to their ability to process temporal data. \cite{bahdanau2015neural} introduced an extension of encoder-decoder architecture to learn alignment and translation. \cite{johnson2017googles} proposed an LSTM-based neural machine translation model to achieve translation between multiple languages in a simple solution. Recurrent neural networks such as LSTM and GRU can theoretically propagate both contextual and sequential information. There have been currently some ERC works modeling the context of discourse based on RNNs. DialogueRNN~\cite[]{majumder2019dialoguernn} updates the status of the speaker and the global information of the conversation by employing multiple GRUs. DialogueCRN~\cite[]{hu2021dialoguecrn} was a cognitive theory-inspired approach that designed a cognitive inference module by exploiting LSTM to capture emotional cues contained in the context.

\subsubsection{Multi-head attention network}
Multi-head ATtention network (MAT) was first proposed by \cite{vaswani2017attention}. It was powerful in feature dependency extraction, leading to remarkable achievements in many tasks. Contrary to RNNs which focus on local information, MAT can extract long-distance elemental dependencies. In recent years, MAT has been widely used in many research areas, such as automatic speech recognition~\cite[]{zhang2020transformer}, natural language processing~\cite[]{vaswani2017attention}, and computer vision~\cite[]{dosovitskiy2021image}. In addition, there exist some pre-trained models constructed with the help of MAT, such as BERT~\cite[]{devlin2019bert}, RoBERTa~\cite[]{liu2020roberta}, and BART~\cite[]{lewis2020bart}. Assuming that the context and speaker information of the utterance is not considered, ERC can be regarded as a text classification task. In this case, each utterance can be fine-tuned with a pre-trained model to extract utterance-level feature. HiTrans~\cite[]{li2020hitrans} adopted BERT to extract utterance-level features, while COSMIC~\cite[]{ghosal2020cosmic} leveraged RoBERTa as a feature extractor for each utterance. In this paper, we follow COSMIC's manner and extract utterance-level features by utilizing RoBERTa.

\section{Methodology}\label{methodology}
\begin{figure*}[htbp]
    \centering
    \includegraphics[width=6.0in]{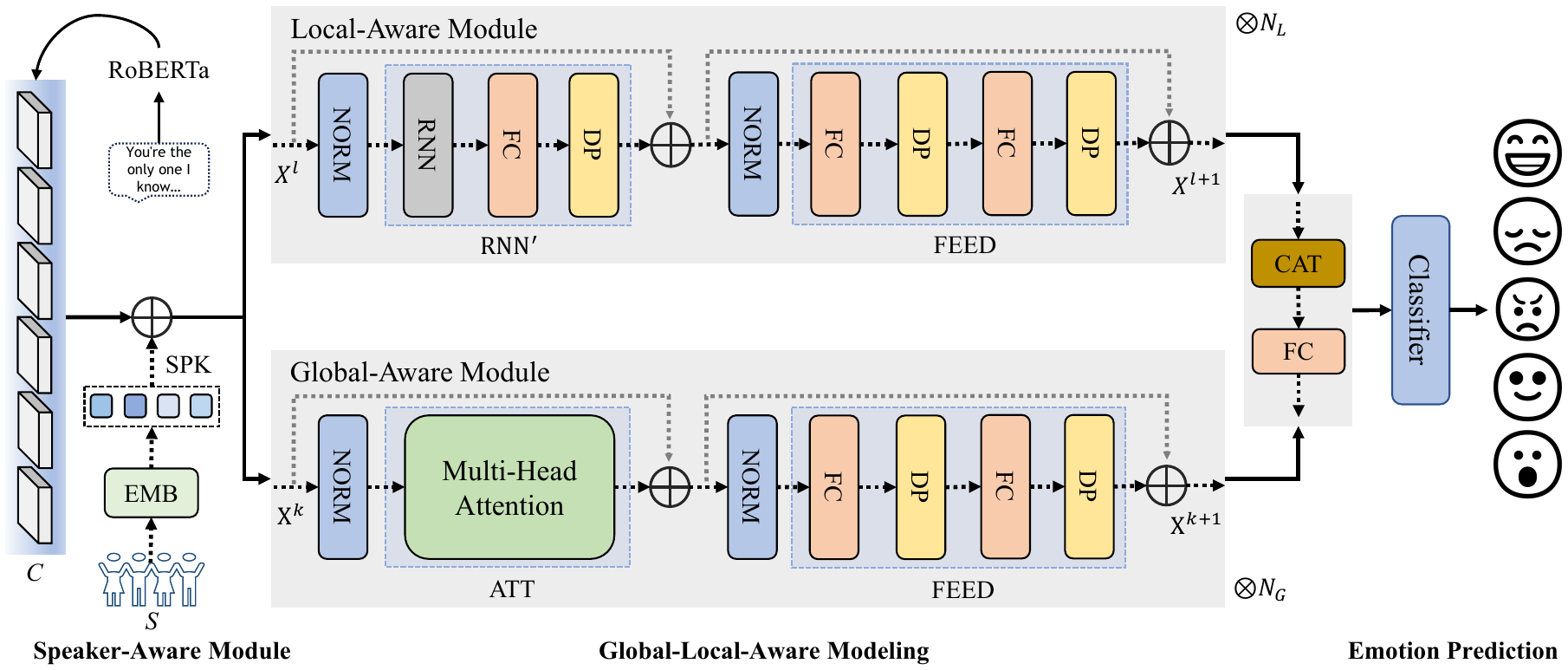}
    \caption{The overall architecture of our proposed DualRAN. Here, $\mathtt{NORM}$, $\mathtt{FC}$, and $\mathtt{DP}$ denote the normalization, fully connected, and dropout layers, respectively; $\mathtt{RNN}$ and $\mathtt{ATT}$ are the single-layer recurrent neural network (e.g., LSTM or GRU) and multi-head attention network; $\mathtt{EMB}$ and $\mathtt{CAT}$ indicate the word embedding and concatenation operations; $N_L$ and $N_G$ denote the number of network layers.}
    \label{fig:overall}
\end{figure*}

We elaborate the proposed dual-stream network structure and its single-stream variants in this section. Our DualRAN is designed with the original intention of combining recurrence-based and attention-based methods to extract both local contextual information and global contextual information. As shown in Figure~\ref{fig:overall}, our DualRAN mainly consists of speaker-aware module, global--local-aware modeling, and emotion prediction. Among them, global--local-aware modeling includes RNN-based local-aware module and MAT-based global-aware module.

\subsection{Task definition}\label{definition}
There exists a conversation $U$ which contains $|U|$ utterances $(u_1,u_2,\ldots,u_{|U|})$ and the corresponding speaker sequence $(s_1,s_2,\ldots,s_{|U|})$, where each utterance $u_i$ corresponds to a speaker $s_i$. The utterances $u_i$ and $u_j$ may be spoken by the same speaker (i.e., $s_i=s_j$) or different speakers (i.e., $s_i\neq s_j$). The task of ERC is to infer the emotion state $e_i$ corresponding to the utterance $u_i$ based on the conversation content and speaker information. The emotion categories in distinct datasets may vary. For instance, in the IEMOCAP dataset, the emotion categories include \textit{happy}, \textit{sad}, \textit{neutral}, \textit{angry}, \textit{excited}, and \textit{frustrated}; while in the MELD dataset, the emotion categories include \textit{joy}, \textit{anger}, \textit{fear}, \textit{disgust}, \textit{sadness}, \textit{surprise}, and \textit{neutral}. Table~\ref{tab:symbol} shows the symbols and their definitions mentioned in this paper.
\begin{table}[htbp]
    \centering
    \renewcommand{\arraystretch}{1.0}
    \setlength{\tabcolsep}{5pt}
    \scriptsize
    \caption{Partial symbols and their definitions.}
    \begin{tabular}{l|l}
    \hline
    Symbols &Definitions \\
	\hline
    $U$  &A conversation or a sequence of utterances \\
    $u_i$ &The $i$-th utterance in that conversation $U$ \\
    $C$ &The utterance-level feature matrix of $U$ \\
	$S$ &The speaker sequence \\
    $s_i$ &The speaker corresponding to the $i$-th utterance \\
    $\mathrm{SPK}$ &The speaker embedding matrix \\
    $E$ & The set of emotion \\
    $e_i$ &The emotion corresponding to the $i$-th utterance \\
    $\mathcal{X}$ & The input to the global--local-aware network \\
    $N_L$ &The number of network layers for the local-aware module \\
    $N_H$ &The number of heads for the multi-head attention network \\
    $N_G$ &The number of network layers for the global-aware module \\
    \hline
    \end{tabular}
    \label{tab:symbol}
\end{table}

\subsection{Speaker-aware module}\label{speaker}
Differences in the identity of speakers may have different effects on the semantics of utterances. To put it another way, the current emotional state of a speaker is influenced not only by his or her own historical utterances but also by the historical utterances of other speakers. That is, there is emotional inertia and emotional contagion within and between speakers. In order to distinguish the influence of different speakers, we add the corresponding identity of the speaker to each utterance, thus implementing speaker-aware encoding. Specifically, we first encode word embedding for each speaker, then add the encoded speaker embedding to the utterance feature, and finally take the obtained new utterance feature as the input to the global--local-aware network. The above process can be formulated as follows:
\begin{equation}
    \label{eq:speaker}
    \begin{split}
    &\mathrm{SPK}=\mathtt{EMB}(S),\\
    &\mathcal{X} = C + \mathrm{SPK},
    \end{split}
\end{equation}
where $S$ denotes the speaker sequence corresponding to the utterance set $U$, while $\mathtt{EMB}$ denotes the word embedding network; $C$ denotes the utterance-level feature matrix of $U$, which is extracted by the method of COSMIC~\cite[]{ghosal2020cosmic}.

\subsection{Global--local-aware modeling}\label{global_local}
The network structure of global--local-aware modeling is simple and effective, as the name suggests, it mainly consists of the local-aware module and global-aware module, which extract local contextual information and global contextual information, respectively. When performing backpropagation, the designed local-aware module and global-aware module are trained simultaneously to update the network parameters. In the following two parts, we describe their network structures respectively.

\subsubsection{Local-aware module}\label{local}
Numerous previous works have demonstrated that modeling the context of utterance is crucial for ERC. Therefore, we construct a local-aware module with a modified RNN. First, in order to extract temporal information of the utterance, we input the utterance feature to the vanilla RNN; then, inspired by the Transformer architecture, we adopt skip connection, i.e., the input and output of RNN are summed; finally, to enhance the expressiveness and stability of the network, we add a feedforward network layer consisting of two fully connected layers. The network structure of the local-aware module can be described by the following equation:
\begin{equation}
    \label{eq:local}
    \begin{split}
    &X_{rnn}^l = \mathtt{NORM}(X^l + \mathtt{RNN^\prime}(X^l)),\\
    &X^{l+1} = \mathtt{NORM}(X_{rnn}^l + \mathtt{FEED}(X_{rnn}^l)),
    \end{split}
\end{equation}
where $X^l$ indicates the $l$-th layer feature matrix composed of all utterances, $l=0,1,\ldots,N_L-1$, and $X^0 = \mathcal{X}$; $\mathtt{NORM}(\cdot)$ denotes the normalization function, and the layer normalization operation is used in our experiments. $\mathtt{RNN^\prime}(\cdot)$ stands for the RNN layer with the addition of a fully connected layer, which can be formulated as,
\begin{equation}
    \label{eq:rnn}
    \mathtt{RNN^\prime}(X^l) = \mathtt{DP}(\mathtt{FC}(\mathtt{RNN}(X^l))),
\end{equation}
$\mathtt{RNN}(\cdot)$ denotes the bidirectional vanilla RNN such as LSTM and GRU; $\mathtt{FC}(\cdot)$ means the fully connected layer, converting the feature dimension of the output to half of the input; $\mathtt{DP}(\cdot)$ indicates the dropout operation. $\mathtt{FEED}(\cdot)$ is the feedforward network layer, which can be expressed as,
\begin{equation}
    \label{eq:feedforward}
    \mathtt{FEED}(X_{rnn}^l) = \mathtt{DP}(\mathtt{FC}(\mathtt{DP}(\alpha(\mathtt{FC}(X_{rnn}^l))))),
\end{equation}
$\alpha(\cdot)$ denotes the activation function, e.g., ReLU. In our experiments, we place $\mathtt{NORM}(\cdot)$ in front of $\mathtt{RNN^\prime}(\cdot)$, i.e.,
\begin{equation}
    \label{eq:localfirst}
    \begin{split}
    &X_{rnn}^l = X^l + \mathtt{RNN^\prime}(\mathtt{NORM}(X^l)),\\
    &X^{l+1} = X_{rnn}^l + \mathtt{FEED}(\mathtt{NORM}(X_{rnn}^l)).
    \end{split}
\end{equation}

\subsubsection{Global-aware module}\label{global}
The local-aware module possesses powerful temporal extraction capability, but it tends to capture local contextual information, while it is quite difficult to aggregate long-distance information. Therefore, we build a global-aware module with the help of Multi-head ATtention network (MAT) to capture global contextual information. Our global-aware module borrows the encoder structure of Transformer, and note that we do not incorporate position encoding because the local-aware module can capture temporal information in the conversation. The network structure of the global-aware module can be expressed as:
\begin{equation}
    \label{eq:global}
    \begin{split}
    &\mathrm{X}_{att}^k = \mathtt{NORM}(\mathrm{X}^k + \mathtt{ATT}(\mathrm{X}^k)),\\
    &\mathrm{X}^{k+1} = \mathtt{NORM}(\mathrm{X}_{att}^k + \mathtt{FEED}(\mathrm{X}_{att}^k)),
    \end{split}
\end{equation}
where $\mathrm{X}^k$ denotes the $k$-th layer feature matrix composed of all utterances, $k=0,1,\ldots,N_G-1$, and $\mathrm{X}^0 = \mathcal{X}$; $\mathtt{ATT}(\cdot)$ and $\mathtt{FEED}(\cdot)$ denote the attention network with multi-head setting and feedforward network layer, respectively. As with the local-aware module, $\mathtt{NORM}(\cdot)$ is placed ahead of $\mathtt{ATT}(\cdot)$, that is,
\begin{equation}
    \label{eq:globalfirst}
    \begin{split}
    &\mathrm{X}_{att}^k = \mathrm{X}^k + \mathtt{ATT}(\mathtt{NORM}(\mathrm{X}^k)),\\
    &\mathrm{X}^{k+1} = \mathrm{X}_{att}^k + \mathtt{FEED}(\mathtt{NORM}(\mathrm{X}_{att}^k)).
    \end{split}
\end{equation}

After both local-aware modeling and global-aware modeling, we obtain the feature matrix $X^{N_L}$ with local information and $\mathrm{X}^{N_G}$ with global information, respectively. Finally, to obtain the global--local-aware feature matrix, we concatenate $X^{N_L}$ and $\mathrm{X}^{N_G}$,
\begin{equation}
    \label{eq:global_local}
	\mathrm{X}_{gl} = W_{gl} \mathtt{CAT}(X^{N_L},\mathrm{X}^{N_G}),
\end{equation}
where $W_{gl}$ is the trainable parameter, and $\mathrm{X}_{gl}$ denotes the feature matrix with global--local awareness.

\subsection{Emotion prediction}\label{prediction}
We make the obtained feature matrix $\mathrm{X}_{gl}$ as the input of the emotion prediction module. Specifically, the feature dimension of $\mathrm{X}_{gl}$ is converted to $|E|$ (number of emotions) through a fully connected layer, and thus the predicted emotion $e_i^\prime$ ($e_i^\prime \in E$) is obtained. The process can be formulated as follows:
\begin{equation}
    \label{eq:emotion}
    \begin{split}
    &y_i^\prime = \mathtt{SMAX}(W_{smax} \mathrm{x}_{{gl},i}),\\
    &e_i^\prime = \mathtt{ARGMAX}(y_i^\prime[k]),
    \end{split}
\end{equation}
where $\mathrm{x}_{{gl},i} \in \mathrm{X}_{gl}$, $W_{smax}$ is the learnable parameter, and $\mathtt{ARGMAX}(\cdot)$ denotes the argmax function. 

\subsection{Model training}\label{training}
To learn the network parameters of DualRAN, we define the loss function as follows:
\begin{equation}
    \label{eq:loss}
    \mathcal{L} = - \frac {1}{\sum_{t=1}^{O} o(t)} \sum_{i=1}^{O}\sum_{j=1}^{o(i)} y_{ij} \log y^{\prime}_{ij} + \eta  \Vert W_{all} \Vert,
\end{equation}
where $o(i)$ is the number of utterances of the $i$-th dialog, and $O$ is the number of all dialogs in training set; $y^{\prime}_{ij}$ denotes the probability distribution of predicted emotion label of the $j$-th utterance in the $i$-th dialog, and $y_{ij}$ denotes the ground truth label; $\eta$ is the L2-regularizer weight, and ${W}_{all}$ is the set of all learnable parameters. 

\subsection{SingleRANs}\label{single}
\begin{figure*}[htbp]
    \centering
    \subfloat[SingleRANv1]{\includegraphics[height=1.5in]{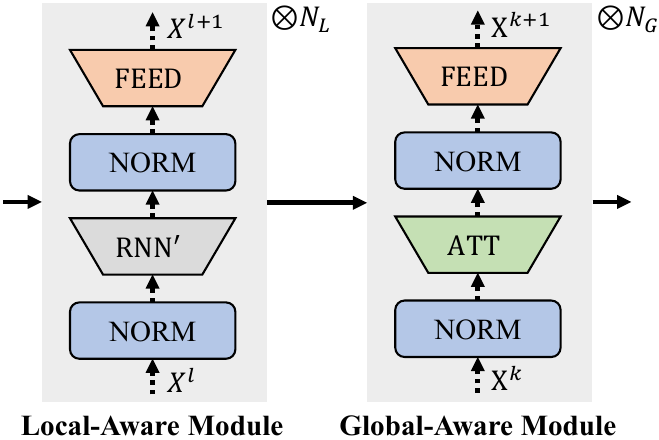}%
    \label{fig:singleranv1}}
    \hfil
    \subfloat[SingleRANv2]{\includegraphics[height=1.5in]{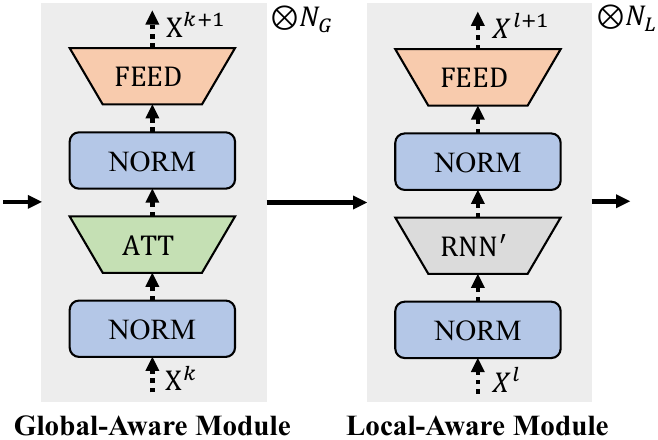}%
    \label{fig:singleranv2}}
    \caption{The network structure of global--local-aware modeling in SingleRAN.}
    \label{fig:singleran}
\end{figure*}

We only change the network structure of global--local-aware modeling in DualRAN to construct the Single-stream Recurrence-Attention Networks (SingleRANs). Like DualRAN, the global--local-aware modeling of SingleRAN contains two modules: local-aware module and global-aware module. The structure of the local-aware module and global-aware module itself remains unchanged, but they are combined in a single-stream and sequential manner, as shown in Figure~\ref{fig:singleran}. According to the order of combining the local-aware module and global-aware module, we divide SingleRAN into two categories, i.e., SingleRANv1 and SingleRANv2.

In SingleRANv1 (see Figure~\ref{fig:singleranv1}), the local-aware module is in the front and the global-aware module is in the back, that is:
\begin{equation}
\label{eq:singlev1}
\begin{split}
    &X^{{l+1}} = \mathtt{LAM}(X^{l}),\\
    &\mathrm{X}^{k+1}= \mathtt{GAM}(\mathrm{X}^{k}),
\end{split}
\end{equation}
where $\mathtt{LAM}(\cdot)$ and $\mathtt{GAM}(\cdot)$ denote the local-aware module and global-aware module, respectively. After the local-aware module of $N_L$ layers, we obtain the feature matrix $X^{N_L}$. Here, $X^{N_L}$ is the output of the local-aware module and is also treated as the input to the global-aware module, i.e., $\mathrm{X}^0 = X^{N_L}$. We obtain the feature matrix $\mathrm{X}_{gl}$ after the global-aware module of $N_G$ layers, which is treated as the input to the prediction module.

In SingleRANv2 (see Figure~\ref{fig:singleranv2}), the global-aware module is in front and the local-aware module is in the back, that is:
\begin{equation}
\label{eq:singlev2}
\begin{split}
    &\mathrm{X}^{k+1} = \mathtt{GAM}(\mathrm{X}^{k}),\\
    &X^{l+1} = \mathtt{LAM}(X^{l}).
\end{split}
\end{equation}
After the global-aware module of $N_G$ layers, we obtain the feature matrix $\mathrm{X}^{N_G}$. Here, $\mathrm{X}^{N_G}$ is the output of the global-aware module and is also treated as the input to the local-aware module, i.e., $X^0 = \mathrm{X}^{N_G}$. Similar to SingleRANv1, after being processed sequentially by the global-aware module of $N_G$ layers and local-aware module of $N_L$ layers, We obtain the feature matrix $\mathrm{X}_{gl}$, and it is treated as the input to the emotion prediction module.

\section{Experimental settings}\label{settings}
\subsection{Datasets}
In order to evaluate the validity of our model, we conduct abundant experiments on four benchmark emotion datasets. These datasets include IEMOCAP\footnote{https://sail.usc.edu/iemocap/}~\cite[]{busso2008iemocap}, MELD\footnote{https://github.com/SenticNet/MELD}~\cite[]{poria2018meld}, EmoryNLP\footnote{https://github.com/emorynlp/emotion-detection}~\cite[]{zahiri2018emotion}, and DailyDialog\footnote{http://yanran.li/dailydialog}~\cite[]{li2017dailydialog}. The statistics of these datasets are reported in Table~\ref{tab:statistics}, from which details of the data splitting can be observed.
\begin{table}[htbp]
    \centering
    \renewcommand{\arraystretch}{1.0}
    \setlength{\tabcolsep}{4pt}
    \scriptsize
    \caption{The statistics of these four datasets used in this Work. \#Dialog and \#Utter denote the number of dialogs and utterances, respectively.}
    \begin{tabular}{cc|cccc}
    \hline
    \multicolumn{2}{c|}{Datasets} &IEMOCAP &MELD &EmoryNLP &DailyDialog\\
    \hline
    \multirow{3}{*}{\#Dialog} &Train &108 &1039 &659  &11118 \\
     &Val &12 &114 &89 &1000 \\
     &Test &31 &280 &79 &1000 \\
    \hline
    \multirow{3}{*}{\#Utter} &Train &5163 &9989 &7551  &87170 \\
     &Val &647 &1109 &954 &8069 \\
     &Test &1623 &2610 &984 &7740 \\
    \hline
    \end{tabular}
    \label{tab:statistics}
\end{table}

\textbf{IEMOCAP} is a dyadic conversational dataset containing 10 unique speakers, of which the first 8 speakers belong to the training set and the last two to the test set. The dataset consists of approximately 12 hours of multimodal dialog data, and we employ only text modality in this work. The dataset contains 152 conversations with a total of 7433 utterances, where these utterances are annotated with one of six emotions, namely \textit{happy}, \textit{sad}, \textit{neutral}, \textit{anger}, \textit{excited}, and \textit{frustrated}. 
\textbf{MELD} is a multi-party multimodal dialog dataset from the TV show "Friends", and we use only text modality in this work. The dataset contains 1433 dialogs with a total of 13708 utterances and has seven emotion categories: \textit{neutral}, \textit{surprise}, \textit{fear}, \textit{sadness}, \textit{joy}, \textit{disgust}, and \textit{anger}. The utterances are labeled with sentiment categories, i.e., \textit{positive}, \textit{negative}, or \textit{neutral}, in addition to being labeled as emotions. 
\textbf{EmoryNLP} collects multi-party conversations from the TV show "Friends". However, the selection of scenes and emotion labels differs from MELD. The dataset contains 827 dialogs with a total of 9489 utterances and seven emotional categories: \textit{sad}, \textit{mad}, \textit{scared}, \textit{powerful}, \textit{peaceful}, \textit{joyful}, and \textit{neutral}. 
\textbf{DailyDialog} is a large scale multi-turn dyadic dialog dataset with the conversations reflecting various topics in daily life. The dataset contains 13118 conversations with a total of 102979 utterances and seven emotion categories: \textit{neutral}, \textit{happiness}, \textit{surprise}, \textit{sadness}, \textit{anger}, \textit{disgust}, and \textit{fear}. The dataset suffers from a severe class imbalance, with over 83\% of the emotion labels being \textit{neutral}.

\begin{figure*}[htbp]
    \centering
    \subfloat[IEMOCAP Dataset]{\includegraphics[width=1.5in]{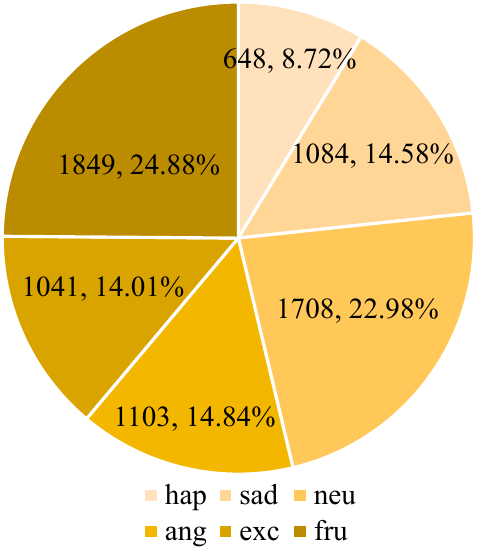}%
    \label{fig:dataiemocap}}
    \hfil
    \subfloat[MELD Dataset]{\includegraphics[width=1.5in]{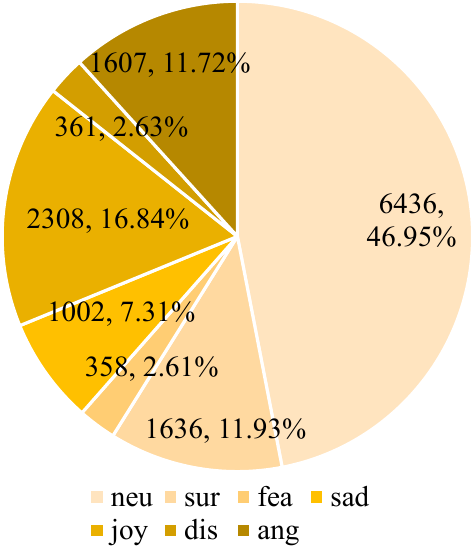}%
    \label{fig:datameld}}
	\hfil
	\subfloat[EmoryNLP Dataset]{\includegraphics[width=1.5in]{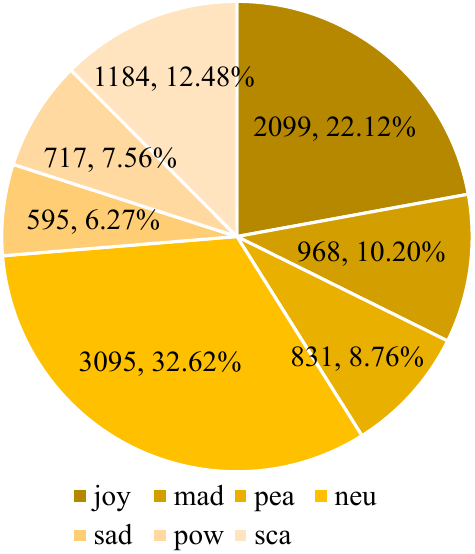}%
    \label{fig:dataemorynlp}}
    \hfil
    \subfloat[DailyDialog Dataset]{\includegraphics[width=1.5in]{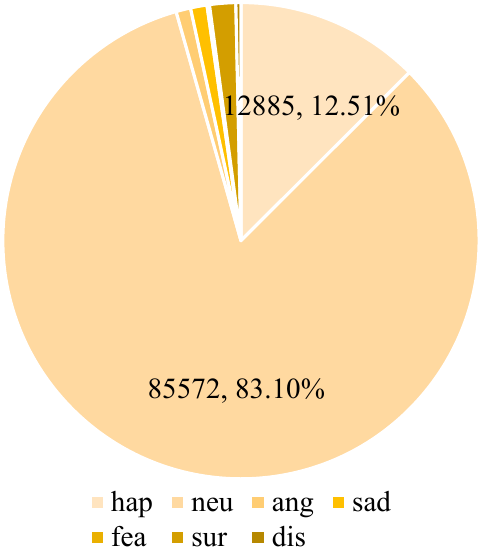}%
    \label{fig:datadailydialog}}
    \caption{Percentage of each emotion in the dataset. Here, \textit{hap} and \textit{neu} denote the abbreviations of \textit{happy} and \textit{neutral}, respectively, and other emotions by analogy.}
    \label{fig:data}
\end{figure*}
Figure~\ref{fig:data} shows the percentage of each emotion in the four datasets. We can observe that all datasets exist the class-imbalanced problem, with DailyDialog being the most serious because of \textit{neutral} accounting for 83.10\%, which poses a high challenge to the ERC model. According to COSMIC\footnote{https://github.com/declare-lab/conv-emotion/tree/master/COSMIC}~\cite[]{ghosal2020cosmic}, we use utterance-level text features which are fine-tuned adopting RoBERTa~\cite[]{liu2020roberta} to implement the ERC task.

\subsection{Baselines and evaluation metrics}
Baselines: The baseline used as comparisons in this work include COSMIC~\cite[]{ghosal2020cosmic}, HiTrans~\cite[]{li2020hitrans}, AGHMN~\cite[]{jiao2020real}, DialogueCRN~\cite[]{hu2021dialoguecrn}, SKAIG-ERC~\cite[]{li2021pastpresent}, DialogXL~\cite[]{shen2021dialogxl}, I-GCN~\cite[]{nie2022igcn}, LR-GCN~\cite[]{ren2022lrgcn}, CauAIN~\cite[]{zhao2022cauain}, CoG-BART~\cite[]{li2022contrast}, and GAR-Net~\cite[]{xu2022gar}. Of all these baselines, COSMIC, SKAIG-ERC, and CauAIN use commonsense knowledge, while the others do not.

\textbf{COSMIC} was a classic work that introduces external commonsense knowledge into emotion recognition in conversation, which leveraged multiple GRUs to integrate commonsense knowledge and extract complex interaction patterns. \textbf{HiTrans} extracted multidimensional contextual information with the use of two hierarchical Transformers and then captured speaker-aware information utilizing pairwise utterance speaker verification. \textbf{AGHMN} constructed the hierarchical memory network and attention gated recurrent units through multiple GRUs respectively to adequately model the context of the utterance. \textbf{DialogueCRN} employed multiple LSTMs to construct the perception phase module and cognition phase module respectively in order to simulate human cognitive behavior, thus enhancing the ability to extract and integrate emotional cues. \textbf{SKAIG-ERC} modeled the context of utterance adopting graph structure and commonsense knowledge, simulating the mental state of the speaker, and then it employed graph convolutional networks for information propagation, which enhanced the emotional representation of the utterance. \textbf{DialogXL} was a pioneering work that applied XLNet to emotion recognition in conversation, which focused on improvements to the recurrence and attention mechanisms to model conversational emotion data. \textbf{I-GCN} was a dialog emotion recognition method that modeled the dialog as a graph structure, and it extracted the semantic correlation information of utterances and temporal sequence information of the conversation with the help of graph convolutional networks. \textbf{LR-GCN} mainly consisted of two modules, latent relation exploration and information propagation, which adopted a multi-branch graph architecture in order to simultaneously capture the speaker information, contextual information of the utterance, and potential correlations between utterances. \textbf{CauAIN} was a cause-aware interaction based model that explicitly modeled speaker dependencies and contextual dependencies of utterance in combination with commonsense knowledge. \textbf{CoG-BART} was an approach that employed both contrast learning and generative models, which can capture the information of long-distance utterances while alleviating the problem of similar emotion. \textbf{GAR-Net} considered both word-level and utterance-level contexts, which constructed an utterance-level graph reasoning network by treating the entire conversation as a fully connected graph.

Evaluation Metrics: Our evaluation metrics include accuracy (\%), weighted F1 (\%), micro F1 (\%), and macro F1 (\%) scores. For the IEMOCAP and MELD datasets, we use accuracy and weighted F1 scores as evaluation metrics; for the EmoryNLP dataset, micro F1 and weighted F1 scores are taken as evaluation metrics; for the DailyDialog dataset, since \textit{neutral} accounts for about 83\% of the dataset, we adopt micro F1 score without \textit{neutral} and macro F1 score to evaluate our model.

\subsection{Training details}
Our experiments are conducted on a single NVIDIA GeForce RTX 3090 and trained in an end-to-end fashion. The deep learning framework which we use is Pytorch with version 2.0.0, and the operating system is Ubuntu 20.04. We choose AdamW~\cite[]{loshchilov2018decoupled} as the optimizer, the L2 regularization factor is 3e-4, and the maximum epochs are set to 100. Other hyperparameter settings for different datasets are displayed in Table~\ref{tab:hyperparameter}. In our experiments, we utilize LSTM~\cite[]{hochreiter1997long} as recurrent neural network for the local-aware module.
\begin{table*}[htbp]
    \centering
    \renewcommand{\arraystretch}{1.0}
    \setlength{\tabcolsep}{10pt}
    \caption{Partial hyperparameter settings for distinct datasets. Here, $N_L$ and $N_G$ indicate the number of layers for the local-aware module and global-aware module, respectively; $N_H$ is the number of heads in the multi-head attention network.}
    \begin{tabular}{c|cccccc}
    \hline
    Datasets &Learning Rate &Batch Size &$N_L$ &$N_H$ &$N_G$ &Dropout Rate\\
	\hline
    IEMOCAP &4e-5 &32 &4 &4 &5 &0.1  \\
    MELD &2e-5 &64 &5 &4 &8 &0.2  \\
    EmoryNLP &1e-5 &128 &5 &4 &5 &0.2  \\
	DailyDialog &2e-5 &128 &5 &4 &6 &0.2  \\
    \hline
    \end{tabular}
    \label{tab:hyperparameter}
\end{table*}

\section{Experimental results and analysis}\label{results}

\subsection{Comparison with baselines}
\begin{table*}[htbp]
    \centering
    \renewcommand{\arraystretch}{1.0}
    \footnotesize
    \setlength{\tabcolsep}{5pt}
    \caption{Performance comparison of our DualRAN with baselines. Results for all baselines are obtained from the original paper. The best results are highlighted in bold. The marker $^\dagger$ indicates the best result from the five experiments, and the marker $^\ddagger$ indicates the confidence interval.}
    \begin{tabular}{c|cc|cc|cc|cc}
    \hline
    \multirow{2}{*}{Methods} &\multicolumn{2}{c|}{IEMOCAP} &\multicolumn{2}{c|}{MELD} &\multicolumn{2}{c|}{EmoryNLP} &\multicolumn{2}{c}{DailyDialog}\\
    \cline{2-9}
           &accuracy &weighted-F1 &accuracy &weighted-F1 &micro-F1 &weighted-F1 &micro-F1 &macro-F1\\ 
    \hline
    COSMIC &- &65.28 &- &65.21 &- &38.11  &58.48 &51.05  \\
    HiTrans &- &64.50 &- &61.94 &- &36.75 &- &-  \\
    AGHMN &63.50 &63.50 &60.30 &58.10 &- &- &- &-  \\
    DialogueCRN &66.05 &66.20 &60.73 &58.39 &- &- &- &-  \\
    SKAIG-ERC &- &66.96 &- &65.18 &- &38.88  &59.75&51.95  \\
    DialogXL &- &65.94 &- &62.41 &- &34.73 &54.93 &-  \\
	I-GCN &65.50 &65.40 &- &60.80 &- &- &- &-  \\
	LR-GCN &68.50 &68.30 &- &65.60 &- &- &- &-  \\
    CauAIN &- &67.61 &- &65.46 &- &-  &58.21 &\textbf{53.85}  \\
    CoG-BART &- &66.18 &- &64.81 &42.58 &39.04 &56.29 &-  \\
    GAR-Net &-  &67.41 &- &62.11 &- &- &56.97 &45.81  \\
	\hline
    \multirow{2}{*}{DualRAN} &\textbf{69.62}$^\dagger$ &\textbf{69.73}$^\dagger$ &\textbf{67.70}$^\dagger$ &\textbf{66.24}$^\dagger$ &\textbf{44.82}$^\dagger$ &\textbf{39.22}$^\dagger$ &\textbf{60.07}$^\dagger$ &{52.89}$^\dagger$  \\
        &69.18$\pm$0.37$^\ddagger$ &69.17$\pm$0.41$^\ddagger$ &67.32$\pm$0.31$^\ddagger$ &66.07$\pm$0.16$^\ddagger$ &44.23$\pm$0.51$^\ddagger$ &39.18$\pm$0.26$^\ddagger$ &59.77$\pm$0.36$^\ddagger$ &52.44$\pm$0.37$^\ddagger$ \\
    \hline
    \end{tabular}
    \label{tab:overallcomparison}
\end{table*}
We report the results of comparative experiments on four emotion datasets in Table~\ref{tab:overallcomparison}, which allow the following conclusions to be drawn:
\begin{itemize}
	\item Our proposed DualRAN achieves remarkable performance on all four emotion datasets, with the most significant improvements in scores on the IEMOCAP dataset. It indicates that DualRAN can adequately model the context and thus effectively extract both global dependency information and local dependency information.
	\item On the IEMOCAP dataset, DualRAN attains 69.62\% accuracy and 69.73\% weighted F1 score. Compared with DialogueCRN, the accuracy of our method is improved by 3.57\%; compared with CoG-BART, the proposed DualRAN has a 3.55\% improvement in the weight F1 score.
	\item On the MELD dataset, the weight F1 score of our DualRAN is 0.78\% higher than that of CauAIN, reaching 66.24\%. DualRAN achieves an accuracy of 67.70\%, which is a 6.97\% improvement relative to that of DialogueCRN. Without using external knowledge, the weighted F1 score of our method is still 1.03\% higher than that of COSMIC.
	\item On the EmoryNLP dataset, the micro F1 score of the proposed DualRAN is 2.24\% higher than that of CoG-BART, achieving 44.82\%. Compared to DialogXL's weighted F1 score, the improvement of our model is 4.49\%, achieving 39.22\%.
	\item On the DailyDialog dataset, DualRAN obtains a micro F1 score of 60.07\%, which is 1.86\% higher than that of CauAIN. The macro F1 score of our model is 0.94\% higher than that of SKAIG-ERC, achieving 52.89\%. However, DualRAN's macro F1 score is 0.96\% lower than CauAIN's and fails to achieve the best performance.
\end{itemize}

Overall, DualRAN shows the most dramatic improvement on the IEMOCAP dataset compared to the results on the other datasets. By examining the dataset, it is found that the number of utterances in a conversation is much higher in the IEMOCAP dataset than in any other dataset. In this case, IEMOCAP relies more on contextual modeling than other datasets. Therefore, our DualRAN shows a definite advantage over other baselines on the IEMOCAP dataset with the help of the global--local-aware network.

\begin{table*}[htbp]
    \centering
    \renewcommand{\arraystretch}{1.0}
    \setlength{\tabcolsep}{2.5pt}
    \caption{Performance comparison of DualRAN with baselines in each emotion. Here, w-F1 denotes the weight F1 score. Results for all baselines are obtained from the original paper.}
    \begin{tabular}{c|cccccc|c|ccccccc|c}
    \hline
    \multicolumn{1}{c|}{\multirow{3}{*}{Methods}} & \multicolumn{7}{c|}{IEMOCAP} & \multicolumn{8}{c}{MELD} \\\cline{2-16}          
    & \textit{hap} & \textit{sad}   & \textit{neu} & \textit{ang} & \textit{exc} & \textit{fru} & \multirow{2}{*}{w-F1} & \textit{neu} & \textit{sur} & \textit{fea} & \textit{sad} & \textit{joy} & \textit{dis} & \textit{ang}  & \multirow{2}{*}{w-F1} \\ \cline{2-7} \cline{9-15}
    & F1 & F1  & F1 & F1 & F1 & F1 &  & F1 & F1 & F1 & F1 & F1 & F1 & F1 &   \\
    \hline
	AGHMN & 52.1  & 73.3 & 58.4 & 61.9 & 69.7 & 62.3 & 63.5 
	& 76.4 & 49.7 & 11.5 & 27.0  & 52.4  & 14.0 & 39.4 & 58.1 \\
	I-GCN & 50.0 & \textbf{83.8} & 59.3 & 64.6 & \textbf{74.3}  & 59.0 & 65.4 
	  & 78.0 & 51.6 & 8.0 & \textbf{38.5} & 54.7 & 11.8 & 43.5 & 60.8 \\
    LR-GCN & 55.5 & 79.1 & 63.8 & \textbf{69.0} & 74.0 & 68.9 & 68.3 
	  & \textbf{80.8} & 57.1 & 0.0 & 36.9 & \textbf{65.8} & 11.0 & 54.7 & 65.6 \\
      \hline
	DualRAN & \textbf{57.81} & 81.17 & \textbf{65.61} & 65.23  & 73.68 & \textbf{69.94} & \textbf{69.73} 
	  & 80.09 & \textbf{58.66} & \textbf{14.71} & 38.39 & 64.59 & \textbf{31.78} & \textbf{54.95} & \textbf{66.24} \\
      \hline
    \end{tabular}%
    \label{tab:eachcomparison}%
\end{table*}%
\begin{figure*}[htbp]
    \centering
    \subfloat[IEMOCAP Dataset]{\includegraphics[width=3.0in]{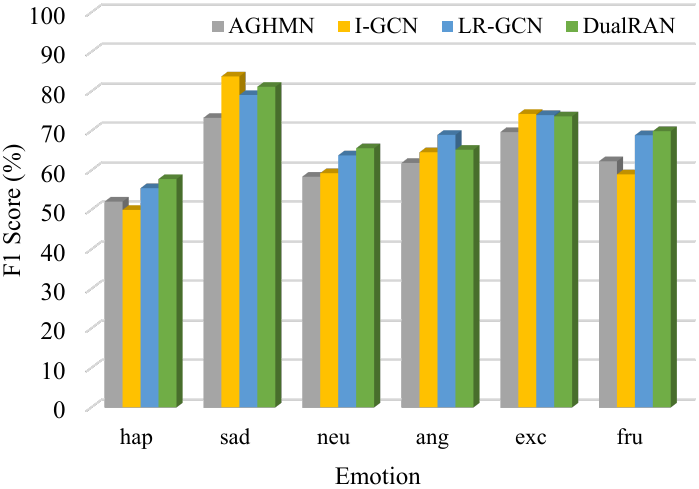}%
    \label{fig:eachiemocap}}
    \hfil
    \subfloat[MELD Dataset]{\includegraphics[width=3.0in]{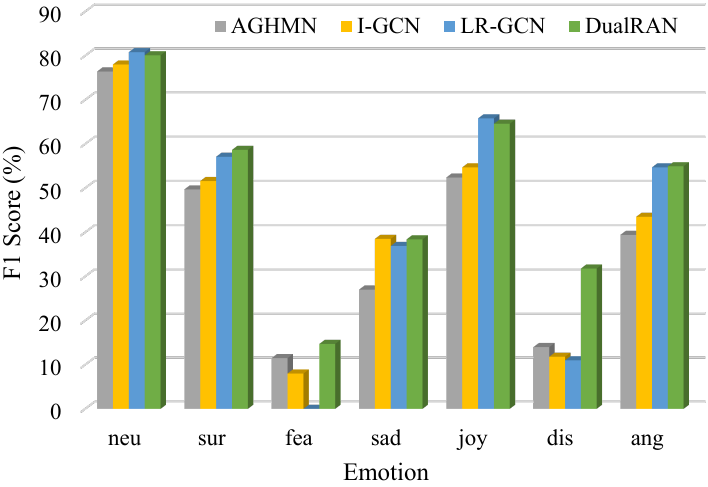}%
    \label{fig:eachmeld}}
    \caption{Comparison of DualRAN with baselines in each emotion on the IEMOCAP and MELD datasets.}
    \label{fig:each}
\end{figure*}
We also record the F1 scores of DualRAN for each emotion on the IEMOCAP and MELD datasets, as shown in Table~\ref{tab:eachcomparison}. It is evident that the proposed DualRAN achieves the best or second best F1 scores for each emotion. For a more intuitive representation, we draw bar charts based on Table~\ref{tab:eachcomparison} to show the comparisons between DualRAN with baselines for each emotion, as shown in Figure~\ref{fig:each}. On the IEMOCAP dataset, DualRAN achieves the best results on \textit{happy}, \textit{neutral}, and \textit{frustrated}, and the second-best F1 scores on \textit{sad}, \textit{angry}, and \textit{excited}, ultimately achieving the best weighted F1 scores. Similar results are obtained on the MELD dataset. Notably, our model achieves 31.78\% F1 scores on \textit{disgust}, an extremely rare emotion category, which is far higher than the other baselines. The above results demonstrate that the proposed DualRAN provides powerful contextual modeling capabilities. In particular, on the MELD dataset, \textit{disgust} can be identified better than other models with the aid of contextual modeling, and the class-imbalanced problem is evaded to some extent.

\subsection{Comparison of SingleRAN with DualRAN}
\begin{table*}[htbp]
    \centering
    \renewcommand{\arraystretch}{1.0}
    \setlength{\tabcolsep}{5pt}
    \caption{Performance comparison of SingleRAN with DualRAN.}
    \begin{tabular}{c|cc|cc|cc|cc}
    \hline
    \multirow{2}{*}{Methods} &\multicolumn{2}{c|}{IEMOCAP} &\multicolumn{2}{c|}{MELD} &\multicolumn{2}{c|}{EmoryNLP} &\multicolumn{2}{c}{DailyDialog}\\
    \cline{2-9}
           &accuracy &weighted-F1 &accuracy &weighted-F1 &micro-F1 &weighted-F1 &micro-F1 &macro-F1\\ 
    \hline 
    DualRAN &\textbf{69.62} &\textbf{69.73} &\textbf{67.70} &\textbf{66.24} &\textbf{44.82} &\textbf{39.22} &\textbf{60.07} &\textbf{52.89}  \\
    \hline
	SingleRANv1 &68.27 &68.41 &67.55 &66.02 &42.78 &38.60  &59.57 &52.14  \\
    SingleRANv2 &68.27 &68.28 &66.36 &65.32 &43.60 &39.22 &59.15 &51.90  \\
	\hline
    \end{tabular}
    \label{tab:singlerancomparison}
\end{table*}
In this subsection, we test the performance of two single-stream variants of DualRAN, namely SingleRANv1 and SingleRANv1, on four datasets. As shown in Table~\ref{tab:singlerancomparison}, on the IEMOCAP dataset, both SingleRANv1 and SingleRANv2 have an accuracy of 68.27\%, which is lower than the score of DualRAN 1.35\%. On the MELD dataset, the weighted F1 score for SingleRANv1 is 0.22\% lower than that of DualRAN, while the weighted F1 score for SingleRANv2 is 0.92\% lower than that of DualRAN. Similar results appear on the EmoryNLP and DailyDialog datasets. The micro F1 score for SingleRANv1 decreases by 2.04\% relative to that for DualRAN on the EmoryNLP dataset, while SingleRANv2's micro F1 score of 43.6\% is 1.22\% lower than DualRAN's. On the DailyDialog dataset, the micro F1 scores for SingleRANv1 and SingleRANv2 declined by 0.5\% and 0.92\% relative to those for DualRAN, respectively. Overall, the performance of two variants, i.e., SingleRANv1 and SingleRANv1, slightly lag behind those of DualRAN.

\subsection{Validity of local- and global-aware modules}
\begin{table*}[htbp]
    \centering
    \renewcommand{\arraystretch}{1.0}
    \setlength{\tabcolsep}{6pt}
    \caption{The validity of global--local-aware modeling on our DualRAN. Here, -w/o L indicates that the local-aware module is removed and only global-aware modeling of the data is performed; -w/o G indicates that the global-aware module is removed and only local-aware modeling of the data is performed.}
    \begin{tabular}{c|cc|cc|cc|cc}
    \hline
    \multirow{2}{*}{Methods} &\multicolumn{2}{c|}{IEMOCAP} &\multicolumn{2}{c|}{MELD} &\multicolumn{2}{c|}{EmoryNLP} &\multicolumn{2}{c}{DailyDialog}\\
    \cline{2-9}
           &accuracy &weighted-F1 &accuracy &weighted-F1 &micro-F1 &weighted-F1 &micro-F1 &macro-F1\\ 
    \hline
    DualRAN &\textbf{69.62} &\textbf{69.73} &\textbf{67.70} &\textbf{66.24} &\textbf{44.82} &\textbf{39.22} &\textbf{60.07} &\textbf{52.89}  \\
    \hline
	-w/o L &64.14 &64.22 &66.63 &65.25 &42.28 &37.12 &58.71 &51.20  \\
	-w/o G &65.06 &65.06 &66.74 &65.74 &42.99 &38.83 &58.26 &51.11  \\
    \hline
    \end{tabular}
    \label{tab:localglobal}
\end{table*}
In this subsection, we remove the local-aware and global-aware modules separately to explore their validity on the performance of DualRAN. From Table~\ref{tab:localglobal}, we can conclude that either removing the local-aware modules or removing the global-aware modules leads to the performance degradation of our model. On the IEMOCAP dataset, the weight F1 score of the model decreases from 69.73\% to 64.22\% when we remove the local-aware module, while that of the model drops to 65.06\% when the global-aware module is removed. The magnitude of reduction suggests that the IEMOCAP dataset is more dependent on local-aware modeling compared to global-aware modeling, and similar patterns are observed for the other datasets (i.e., MELD and EmoryNLP) except for the DailyDialog dataset. Overall, the impact on the IEMOCAP dataset is more significant than that on the others when removing any module. This is due to the fact that a conversation in the IEMOCAP dataset contains more utterances and relies more on contextual modeling than the other datasets.

\subsection{Effect of number of network layers}
\begin{figure}[htbp]
    \centering
    \includegraphics[width=3.0in]{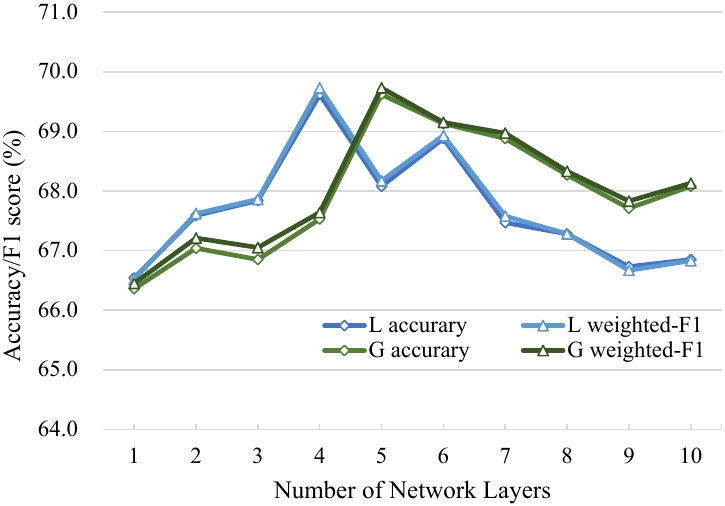}
    \caption{The impact of the number of network layers on the model performance. Results are from experiments on the IEMOCAP dataset. The blue lines indicate the effect of probing the local-aware module (denoted using L) on our model, and the green lines denote the effect of exploring global-aware module (denoted using G) on DualRAN.}
    \label{fig:numlayer}
\end{figure}
To investigate the effect of the number of network layers for global--local-aware modeling on the performance of DualRAN, we conduct ablation studies related to the number of network layers in this subsection. We fix the number of network layers for the global-aware module, while adjusting those for the local-aware module and recording the experimental results. As shown in Figure~\ref{fig:numlayer}, the blue lines depict the effect of the number of network layers for the local-aware module on the accuracy and weight F1 scores. Note that these results are derived from experiments conducted on the IEMOCAP dataset. It can be found that as the number of network layers increases, both the accuracy score and the weight F1 score fluctuate around the optimal performance, i.e., roughly showing an increasing trend followed by a decreasing trend. Similarly, fixing the number of network layers for the local-aware module, we adjust the number of network layers for the global-aware module to explore its impact on the performance of our DualRAN. As shown by green lines in Figure~\ref{fig:numlayer}, the performance of the proposed model tends to increase and then decrease as the number of network layers for the global-aware module increases.

\subsection{Impact of distinct RNNs}
\begin{figure}[htbp]
    \centering
    \includegraphics[width=3.0in]{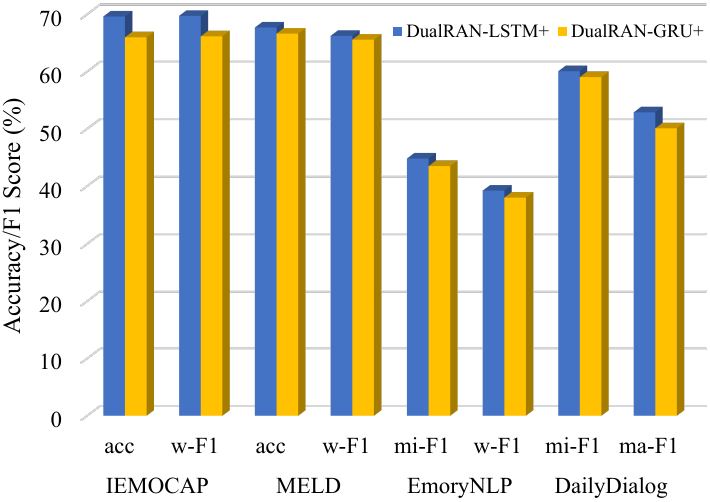}
    \caption{The impact of different RNNs on the performance of DualRAN. DualRAN adopts improved LSTM by default (denoted as DualRAN-LSTM+). DualRAN-GRU+ indicates the use of improved GRU as local-aware module. Here, acc, w-F1, mi-F1, ma-F1 denote accuracy, weighted F1, micro F1, and macro F1 scores, respectively.}
    \label{fig:rnn+}
\end{figure}
\begin{figure}[htbp]
    \centering
    \includegraphics[width=3.0in]{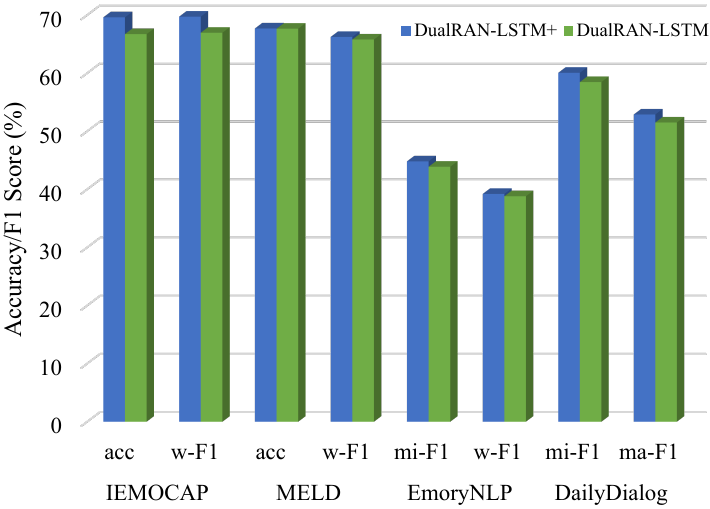}
    \caption{Comparison between the use of improved LSTM and vanilla LSTM. DualRAN-LSTM indicates the direct use of vanilla LSTM without the use of improved LSTM.}
    \label{fig:lstm}
\end{figure}
We test the effect of different RNNs on DualRAN in this subsection. Figure~\ref{fig:rnn+} shows the experimental results using improved LSTM and improved GRU as the local-aware module, respectively. We can see that the accuracy and weight F1 scores by adopting improved LSTM are higher relative to those by adopting improved GRU in all benchmark datasets. On the whole, better results can be obtained with improved LSTM, which indicates that improved LSTM can perform better local-aware modeling relative to improved GRU. Figure~\ref{fig:lstm} shows the comparison between employing improved LSTM and vanilla LSTM. We can reveal that DualRAN utilizing improved LSTM achieves better performance relative to vanilla LSTM on the four datasets. This situation suggests that the inclusion of skip connections and feedforward layers in local-aware module is beneficial in enhancing the expressiveness of the model.

\subsection{Effect of speaker identity}
\begin{figure}[htbp]
    \centering
    \includegraphics[width=3.0in]{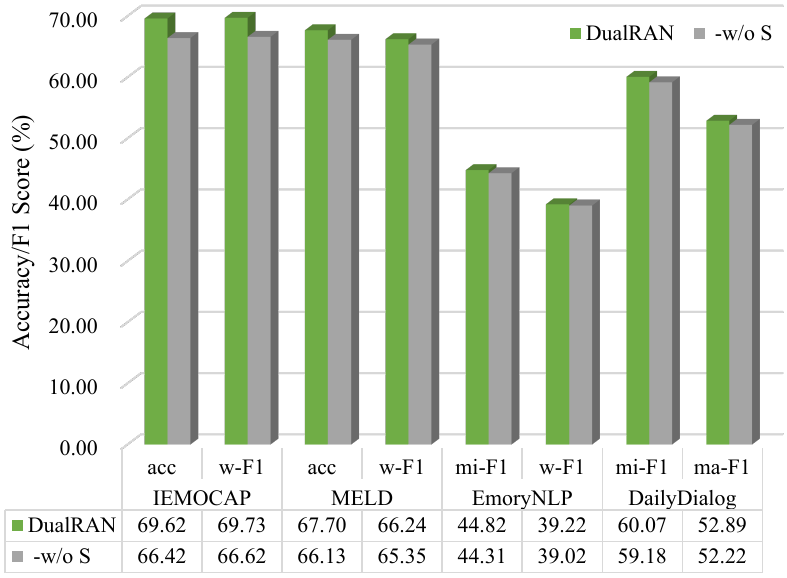}
    \caption{The impact of speaker information on the performance of DualRAN. Here, -w/o S denotes the removal of speaker information.}
    \label{fig:speaker}
\end{figure}
To explore the effect of speaker identity on the proposed DualRAN, we conduct the ablation experiments on speaker information, and the results are displayed in Figure~\ref{fig:speaker}. On the IEMOCAP dataset, the accuracy of our model decreases from 69.62\% to 66.42\% when speaker embedding is not employed, a decrease of 3.20\%. On the MELD dataset, the weight F1 score drops to 65.35\% when speaker information is removed. Similar performance decreases are found on the EmoryNLP and DailyDialog datasets. These phenomena suggest that speaker identity can effectively model emotional inertia and emotional contagion within and between speakers, which is beneficial to improve the performance of the model.

\subsection{Impact of skip connection}
\begin{table*}[htbp]
    \centering
    \renewcommand{\arraystretch}{1.0}
    \setlength{\tabcolsep}{5pt}
    \caption{The impact of skip connection on the proposed DualRAN. Here, -w/o SC-L, -w/o SC-G, and -w/o SC-LG indicate the deletion of skip connections for the local aware module, the global aware module, and both, respectively.}
    \begin{tabular}{c|cc|cc|cc|cc}
    \hline
    \multirow{2}{*}{Methods} &\multicolumn{2}{c|}{IEMOCAP} &\multicolumn{2}{c|}{MELD} &\multicolumn{2}{c|}{EmoryNLP} &\multicolumn{2}{c}{DailyDialog}\\
    \cline{2-9}
           &accuracy &weighted-F1 &accuracy &weighted-F1 &micro-F1 &weighted-F1 &micro-F1 &macro-F1\\ 
    \hline
    DualRAN &\textbf{69.62} &\textbf{69.73} &\textbf{67.70} &\textbf{66.24} &\textbf{44.82} &\textbf{39.22} &\textbf{60.07} &\textbf{52.89}  \\
    \hline
	-w/o SC-L &68.33 &68.45 &66.59 &64.99 &43.60 &37.23 &58.98 &49.52  \\
	-w/o SC-G &66.48 &66.48 &67.09 &65.84 &43.50 &38.58 &59.87 &52.28  \\
	-w/o SC-LG &64.63 &64.82 &64.64 &63.13 &35.16 &33.32 &56.83 &49.70 \\
    \hline
    \end{tabular}
    \label{tab:skip}
\end{table*}
Several studies~\cite[]{he2016deep,li2018deeper,li2021deepgcns} have demonstrated that skip connection can improve the expressiveness and stability of the model, so we add skip connections to both the local-aware module and global-aware module. To demonstrate the effectiveness of skip connection, we conduct the ablation studies on skip connection in this subsection, and the results are depicted in Table~\ref{tab:skip}. As we can observe, the performance of the proposed model appears to degrade on all datasets whether skip connections are removed from the local-aware module or global-aware module. As expected, the degradation of our model is even more pronounced when we remove skip connections of both modules at the same time. These phenomena suggest that introducing skip connections in the global--local-aware network can effectively promote the performance of the model.

\subsection{Results of sentiment classification}
\begin{table*}[htbp]
    \centering
    \renewcommand{\arraystretch}{1.0}
    \setlength{\tabcolsep}{10pt}
    \caption{The statistics of the scheme of emotion merging. Note that the MELD dataset itself contains sentiment labels.}
    \begin{tabular}{c|c|c|c|c}
    \hline
    Sentiments &IEMOCAP &MELD &EmoryNLP &DailyDialog\\
    \hline
	negative &sad, angry, frustrated &negative &mad, sad, scared &anger, sad, fear, disgust \\
	neutral &neutral &neutral &neutral &neutral \\
	positive &happy, excited &positive	&joyful, peaceful, powerful
	&happy, surprise \\
    \hline
    \end{tabular}
    \label{tab:merging}
\end{table*}
In this subsection, we replace emotion with sentiment as the classified target. Accordingly, we transform DualRAN into a tri-classification (i.e., neutral, positive, and negative) model. Note that since the IEMOCAP, EmoryNLP, and DailyDialog datasets contain no sentiment labels, we require to merge the original emotion labels. The specific scheme of merging is shown in Table~\ref{tab:merging}.

\begin{table*}[htbp]
    \centering
    \renewcommand{\arraystretch}{1.0}
    \setlength{\tabcolsep}{4pt}
    \caption{Results of sentiment classification. Here, -emo and -sen denote emotion and sentiment classification, respectively. COSMIC's results are from the original paper.}
    \begin{tabular}{c|cc|cc|cc|cc}
    \hline
    \multirow{2}{*}{Methods} &\multicolumn{2}{c|}{IEMOCAP} &\multicolumn{2}{c|}{MELD} &\multicolumn{2}{c|}{EmoryNLP} &\multicolumn{2}{c}{DailyDialog}\\
    \cline{2-9}
           &accuracy &weighted-F1 &accuracy &weighted-F1 &micro-F1 &weighted-F1 &micro-F1 &macro-F1\\ 
    \hline
	COSMIC-emo &- &65.28 &- &65.21 &- &38.11 &58.48 &51.05  \\
    DualRAN-emo &69.62 &69.73 &67.70 &66.24 &44.82 &39.22 &60.07 &52.89  \\
    \hline
	COSMIC-sen &- &- &- &73.20 &- &56.51 &- &-  \\
	DualRAN-sen &82.38 &82.55 &73.22 &73.21 &57.42 &57.53 &60.66 &66.08  \\
    \hline
    \end{tabular}
    \label{tab:sentiment}
\end{table*}
As shown in Table~\ref{tab:sentiment}, the results of DualRAN are similar to COSMIC after the coarsening of emotion into sentiment, and the performance is improved on all datasets. For instance, the accuracy of DualRAN on the IEMOCAP dataset improves from 69.62\% to 82.38\%, an increase of 12.76\%. Although, relative to COSMIC's results of sentiment classification, the weight F1 scores of our DualRAN on the MELD and EmoryNLP datasets are improved, the enhancements are limited. This situation is mainly due to the fact that most of the models can be easily classified after the fine-grained emotions is coarsened into sentiments. To put it in a nutshell, the dataset becomes relatively simple, so it is not necessary to model both local and global contexts to achieve better results.

\subsection{Case study}
\begin{figure}[htbp]
    \centering
    \includegraphics[width=3.0in]{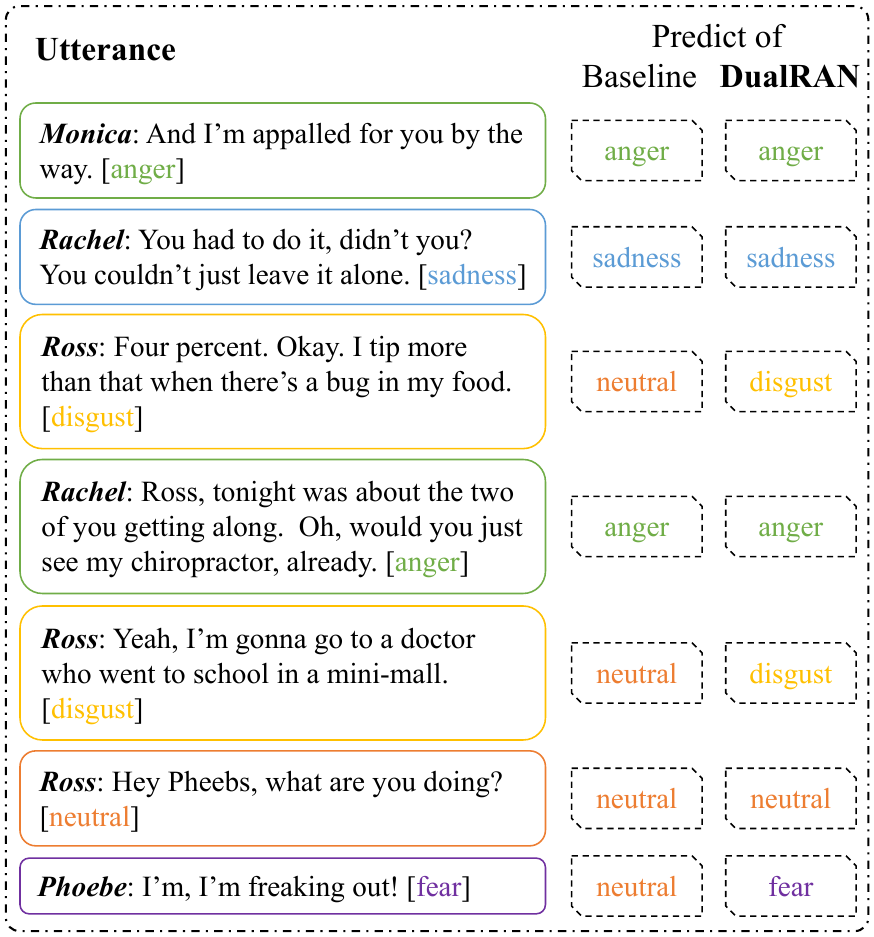}
    \caption{Case study from the MELD dataset. [emotion] at the end of the utterance is the true label.}
    \label{fig:case}
\end{figure}
We extract raw utterances from the MELD dataset for the case study. As shown in Figure~\ref{fig:case}, several baseline models (e.g., LR-GCN) tend to classify the utterances with true labels of \textit{disgust} and \textit{fear} as \textit{neutral}. This is due to the problem of class imbalance in the MELD dataset, where \textit{disgust} and \textit{fear} belong to the minority class, while \textit{neutral} belongs to the majority class. The baseline cannot adequately model the context and tends to predict the emotion of utterance as the majority class, leading to model failure. As can be seen in Figure~\ref{fig:case}, our proposed DualRAN, in contrast to the baseline, takes into account both global and local information and can identify the utterances with true emotions of \textit{disgust} and \textit{fear} as the correct emotions very well. Overall, relative to the baseline models, DualRAN can sufficiently capture local and global contextual information to accurately identify the minority categories by exploiting the global--local-aware network in some scenarios.

\subsection{Limitations}
\begin{figure*}[htbp]
    \centering
    \subfloat[IEMOCAP Dataset]{\includegraphics[width=3.0in]{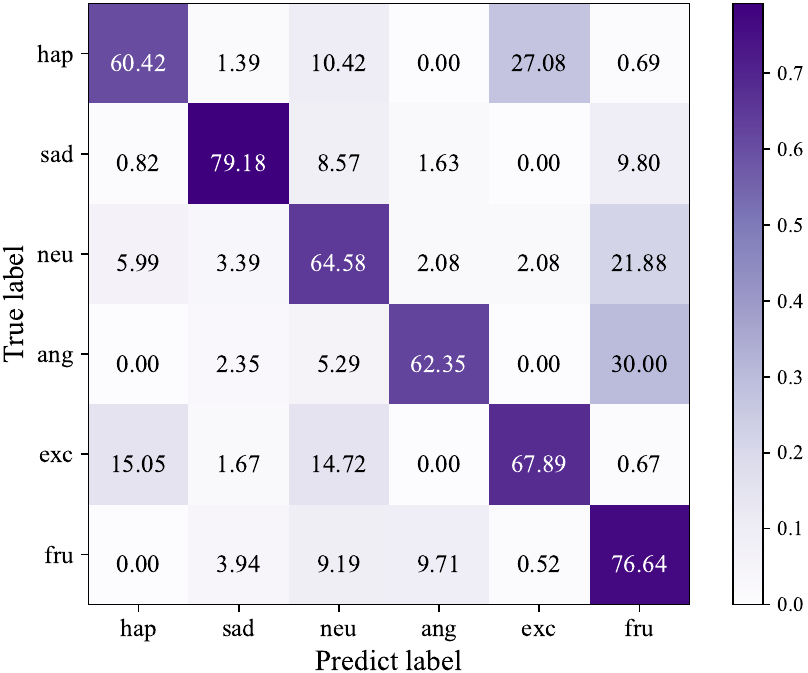}%
    \label{fig:cmiemocap}}
    \hfil
    \subfloat[MELD Dataset]{\includegraphics[width=3.0in]{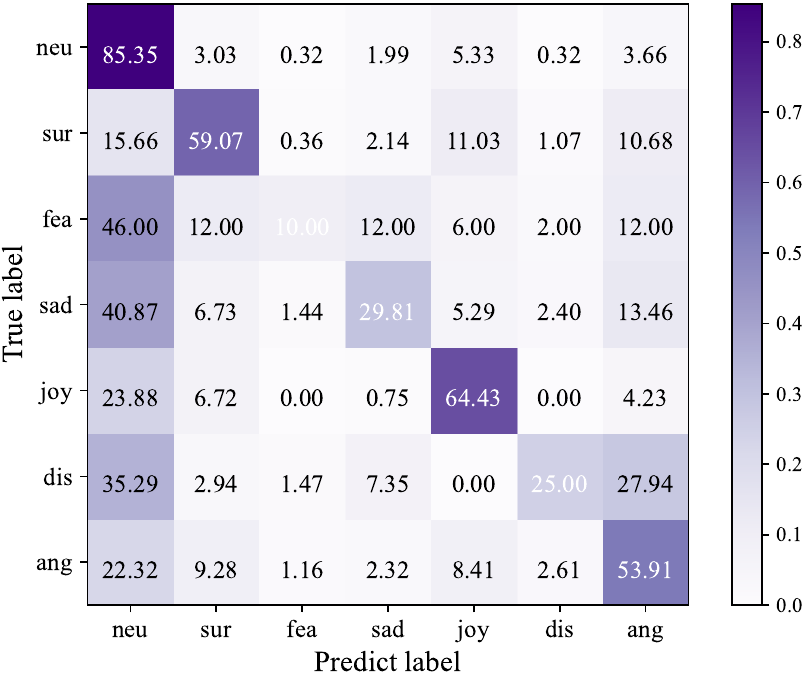}%
    \label{fig:cmmeld}}
	\hfil
	\subfloat[EmoryNLP Dataset]{\includegraphics[width=3.0in]{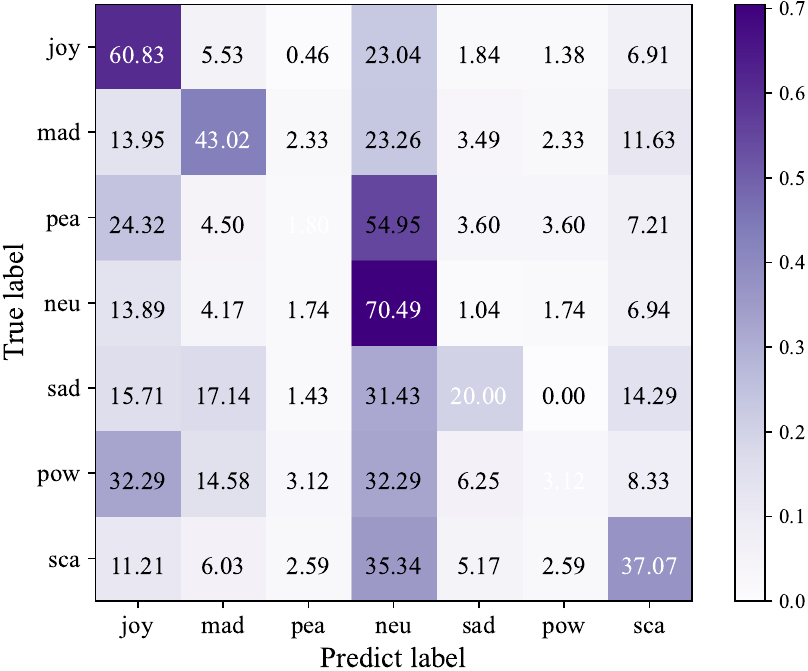}%
    \label{fig:cmemorynlp}}
    \hfil
    \subfloat[DailyDialog Dataset]{\includegraphics[width=3.0in]{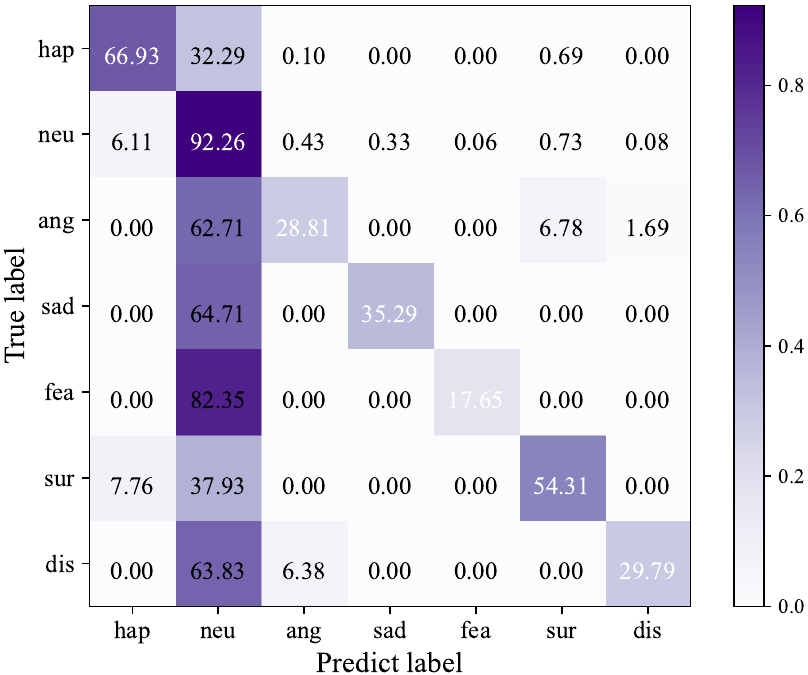}%
    \label{fig:cmdailydialog}}
    \caption{Confusion matrices of the proposed method on the IEMOCAP, MELD, EmoryNLP, and DailyDialog datasets.}
    \label{fig:cm}
\end{figure*}
As shown in Figure~\ref{fig:cm}, we depict the performance of DualRAN on four public emotion datasets with the confusion matrices. It can be seen that the proposed DualRAN achieves superior result on the IEMOCAP dataset. Similar to some previous models, DualRAN works less well on the DailyDialog dataset, as shown in Figure~\ref{fig:cmdailydialog}. One main factor is that the DailyDialog dataset suffers from an extreme class imbalance, i.e., the utterances annotated as \textit{neutral} account for a very large proportion of the dataset, causing DualRAN to be biased toward \textit{neutral} during training. It is evident from Figure~\ref{fig:cmdailydialog} that most utterances tend to be predicted as \textit{neutral}. It is assumed that the performance on the DailyDialog dataset will be improved if \textit{neutral} is removed. The problem of class imbalance is also present in the MELD dataset. As shown in Figure~\ref{fig:cmmeld}, the utterances with true labels of \textit{fear}, \textit{sadness}, and \textit{disgust} incline to be classified as \textit{neutral}. After examining the MELD dataset, it is found that these three emotions belong to the minority class. In addition, DualRAN suffers from the problem of similar emotion, i.e., some utterances are easily misidentified as another similar emotions. For example, as shown in Figure~\ref{fig:cmiemocap}, the utterances whose true labels are \textit{happy} are easily predicted as \textit{excited}, and the utterances with true emotions of \textit{angry} are easily classified as \textit{frustrated} on the IEMOCAP dataset.

\section{Conclusion and prospect}\label{conclusion}
In this paper, we propose a Dual-stream Recurrence-Attention Network (DualRAN) with global--local-aware capability to adequately capture both local and global contextual information of the utterance. The proposed DualRAN is a simple and effective dual-stream network consisting of local- and global-aware modules and focuses on the combination of recurrence-based and attention-based methods. In order to construct the local-aware module, we improve the structure of vanilla RNN referring to Transformer, that is adding the skip connection and feedforward network layer to enhance the expressiveness of the network. To explore the importance of speaker information for the ERC task, we encode the speaker identity and then add it to the corresponding utterance feature. Additionally, based on the local- and global-aware modules of DualRAN, we construct two single-stream recurrence-attention networks, i.e., SingleRANv1 and SingleRANv2. We conduct extensive comparison experiments and the results demonstrate that our proposed model outshines all baselines by an absolute margin. Meanwhile, we perform ablation experiments for each component and the empirical results prove the validity of these components. 

In future research, we will work on addressing the class imbalance problem that is widespread in benchmark emotion datasets. Contrastive learning has gained great achievements in the field of computer vision in recent years, and extending it to the ERC task is a feasible solution to the imbalance problem. Another viable option is to generate some minority class samples with the help of the large language model in order to realize class balancing as much as possible. For the similar emotion problem, pushing away the distance of different class samples with the aid of contrastive learning is a promising scheme. With the potential of multimodal learning, we also intend to further explore the multimodal setting of the ERC task.

\section*{CRediT authorship contribution statement}
\textbf{Jiang Li:} Conceptualization, Methodology, Data curation, Software, Validation, Formal analysis, Investigation, Visualization, Writing - original draft, Writing - review \& editing, Project administration. \textbf{Xiaoping Wang:} Supervision, Writing - review \& editing, Funding acquisition. \textbf{Zhigang Zeng:} Supervision, Funding acquisition.

\section*{Declaration of competing interest}
The authors declare that they have no known competing financial interests or personal relationships that could have appeared to influence the work reported in this paper.

\section*{Data availability}
Data will be made available on request.

\section*{Acknowledgments}
This work was supported in part by the National Natural Science Foundation of China under Grant 62236005, 61936004, and U1913602.

\balance
\bibliographystyle{elsarticle-harv}
\bibliography{dualran.bib}

\begin{thebibliography}{45}
\expandafter\ifx\csname natexlab\endcsname\relax\def\natexlab#1{#1}\fi
\providecommand{\url}[1]{\texttt{#1}}
\providecommand{\href}[2]{#2}
\providecommand{\path}[1]{#1}
\providecommand{\DOIprefix}{doi:}
\providecommand{\ArXivprefix}{arXiv:}
\providecommand{\URLprefix}{URL: }
\providecommand{\Pubmedprefix}{pmid:}
\providecommand{\doi}[1]{\href{http://dx.doi.org/#1}{\path{#1}}}
\providecommand{\Pubmed}[1]{\href{pmid:#1}{\path{#1}}}
\providecommand{\bibinfo}[2]{#2}
\ifx\xfnm\relax \def\xfnm[#1]{\unskip,\space#1}\fi
\bibitem[{Bahdanau et~al.(2015)Bahdanau, Cho and Bengio}]{bahdanau2015neural}
\bibinfo{author}{Bahdanau, D.}, \bibinfo{author}{Cho, K.},
  \bibinfo{author}{Bengio, Y.}, \bibinfo{year}{2015}.
\newblock \bibinfo{title}{Neural machine translation by jointly learning to
  align and translate}, in: \bibinfo{booktitle}{Proceedings of the 3rd
  International Conference on Learning Representations}, pp.
  \bibinfo{pages}{1--15}.
\bibitem[{Busso et~al.(2008)Busso, Bulut, Lee, Kazemzadeh, Mower, Kim, Chang,
  Lee and Narayanan}]{busso2008iemocap}
\bibinfo{author}{Busso, C.}, \bibinfo{author}{Bulut, M.}, \bibinfo{author}{Lee,
  C.C.}, \bibinfo{author}{Kazemzadeh, A.}, \bibinfo{author}{Mower, E.},
  \bibinfo{author}{Kim, S.}, \bibinfo{author}{Chang, J.N.},
  \bibinfo{author}{Lee, S.}, \bibinfo{author}{Narayanan, S.S.},
  \bibinfo{year}{2008}.
\newblock \bibinfo{title}{{IEMOCAP}: interactive emotional dyadic motion
  capture database}.
\newblock \bibinfo{journal}{Language Resources and Evaluation}
  \bibinfo{volume}{42}, \bibinfo{pages}{335--359}.
\newblock \DOIprefix\doi{10.1007/s10579-008-9076-6}.
\bibitem[{Chen and Huang(2021)}]{chen2021anovel}
\bibinfo{author}{Chen, Q.}, \bibinfo{author}{Huang, G.}, \bibinfo{year}{2021}.
\newblock \bibinfo{title}{A novel dual attention-based blstm with hybrid
  features in speech emotion recognition}.
\newblock \bibinfo{journal}{Engineering Applications of Artificial
  Intelligence} \bibinfo{volume}{102}, \bibinfo{pages}{104277}.
\newblock \DOIprefix\doi{10.1016/j.engappai.2021.104277}.
\bibitem[{Chung et~al.(2014)Chung, Gulcehre, Cho and
  Bengio}]{chung2014empirical}
\bibinfo{author}{Chung, J.}, \bibinfo{author}{Gulcehre, C.},
  \bibinfo{author}{Cho, K.}, \bibinfo{author}{Bengio, Y.},
  \bibinfo{year}{2014}.
\newblock \bibinfo{title}{Empirical evaluation of gated recurrent neural
  networks on sequence modeling}, in: \bibinfo{booktitle}{NIPS 2014 Workshop on
  Deep Learning}, pp. \bibinfo{pages}{1--9}.
\bibitem[{Devlin et~al.(2019)Devlin, Chang, Lee and Toutanova}]{devlin2019bert}
\bibinfo{author}{Devlin, J.}, \bibinfo{author}{Chang, M.W.},
  \bibinfo{author}{Lee, K.}, \bibinfo{author}{Toutanova, K.},
  \bibinfo{year}{2019}.
\newblock \bibinfo{title}{{BERT}: Pre-training of deep bidirectional
  transformers for language understanding}, in: \bibinfo{booktitle}{Proceedings
  of the 2019 Conference of the North {A}merican Chapter of the Association for
  Computational Linguistics: Human Language Technologies, Volume 1 (Long and
  Short Papers)}, \bibinfo{publisher}{Association for Computational
  Linguistics}, \bibinfo{address}{Minneapolis, Minnesota}. pp.
  \bibinfo{pages}{4171--4186}.
\newblock \DOIprefix\doi{10.18653/v1/N19-1423}.
\bibitem[{Dosovitskiy et~al.(2021)Dosovitskiy, Beyer, Kolesnikov, Weissenborn,
  Zhai, Unterthiner, Dehghani, Minderer, Heigold, Gelly, Uszkoreit and
  Houlsby}]{dosovitskiy2021image}
\bibinfo{author}{Dosovitskiy, A.}, \bibinfo{author}{Beyer, L.},
  \bibinfo{author}{Kolesnikov, A.}, \bibinfo{author}{Weissenborn, D.},
  \bibinfo{author}{Zhai, X.}, \bibinfo{author}{Unterthiner, T.},
  \bibinfo{author}{Dehghani, M.}, \bibinfo{author}{Minderer, M.},
  \bibinfo{author}{Heigold, G.}, \bibinfo{author}{Gelly, S.},
  \bibinfo{author}{Uszkoreit, J.}, \bibinfo{author}{Houlsby, N.},
  \bibinfo{year}{2021}.
\newblock \bibinfo{title}{An image is worth 16x16 words: Transformers for image
  recognition at scale}, in: \bibinfo{booktitle}{Proceedings of International
  Conference on Learning Representations}, pp. \bibinfo{pages}{1--21}.
\bibitem[{Ghosal et~al.(2020)Ghosal, Majumder, Gelbukh, Mihalcea and
  Poria}]{ghosal2020cosmic}
\bibinfo{author}{Ghosal, D.}, \bibinfo{author}{Majumder, N.},
  \bibinfo{author}{Gelbukh, A.}, \bibinfo{author}{Mihalcea, R.},
  \bibinfo{author}{Poria, S.}, \bibinfo{year}{2020}.
\newblock \bibinfo{title}{Cosmic: Commonsense knowledge for emotion
  identification in conversations}, in: \bibinfo{booktitle}{Findings of the
  Association for Computational Linguistics: EMNLP 2020},
  \bibinfo{publisher}{Association for Computational Linguistics}. pp.
  \bibinfo{pages}{2470--2481}.
\bibitem[{Hazarika et~al.(2020)Hazarika, Zimmermann and
  Poria}]{hazarika2020misa}
\bibinfo{author}{Hazarika, D.}, \bibinfo{author}{Zimmermann, R.},
  \bibinfo{author}{Poria, S.}, \bibinfo{year}{2020}.
\newblock \bibinfo{title}{Misa: Modality-invariant and -specific
  representations for multimodal sentiment analysis}, in:
  \bibinfo{booktitle}{Proceedings of the 28th {ACM} International Conference on
  Multimedia}, \bibinfo{publisher}{{ACM}}. pp. \bibinfo{pages}{1122--1131}.
\newblock \DOIprefix\doi{10.1145/3394171.3413678}.
\bibitem[{He et~al.(2016)He, Zhang, Ren and Sun}]{he2016deep}
\bibinfo{author}{He, K.}, \bibinfo{author}{Zhang, X.}, \bibinfo{author}{Ren,
  S.}, \bibinfo{author}{Sun, J.}, \bibinfo{year}{2016}.
\newblock \bibinfo{title}{Deep residual learning for image recognition}, in:
  \bibinfo{booktitle}{2016 {IEEE} Conference on Computer Vision and Pattern
  Recognition ({CVPR})}, \bibinfo{publisher}{{IEEE}}. pp.
  \bibinfo{pages}{770--778}.
\newblock \DOIprefix\doi{10.1109/cvpr.2016.90}.
\bibitem[{Hochreiter and Schmidhuber(1997)}]{hochreiter1997long}
\bibinfo{author}{Hochreiter, S.}, \bibinfo{author}{Schmidhuber, J.},
  \bibinfo{year}{1997}.
\newblock \bibinfo{title}{Long short-term memory}.
\newblock \bibinfo{journal}{Neural Computation} \bibinfo{volume}{9},
  \bibinfo{pages}{1735--1780}.
\newblock \DOIprefix\doi{10.1162/neco.1997.9.8.1735}.
\bibitem[{Hu et~al.(2021)Hu, Wei and Huai}]{hu2021dialoguecrn}
\bibinfo{author}{Hu, D.}, \bibinfo{author}{Wei, L.}, \bibinfo{author}{Huai,
  X.}, \bibinfo{year}{2021}.
\newblock \bibinfo{title}{Dialoguecrn: Contextual reasoning networks for
  emotion recognition in conversations}, in: \bibinfo{booktitle}{Proceedings of
  the Annual Meeting of the Association for Computational Linguistics and the
  International Joint Conference on Natural Language Processing},
  \bibinfo{publisher}{Association for Computational Linguistics}. pp.
  \bibinfo{pages}{7042--7052}.
\newblock \DOIprefix\doi{10.18653/v1/2021.acl-long.547}.
\bibitem[{Johnson et~al.(2017)Johnson, Schuster, Le, Krikun, Wu, Chen, Thorat,
  Vi{\'e}gas, Wattenberg, Corrado, Hughes and Dean}]{johnson2017googles}
\bibinfo{author}{Johnson, M.}, \bibinfo{author}{Schuster, M.},
  \bibinfo{author}{Le, Q.V.}, \bibinfo{author}{Krikun, M.},
  \bibinfo{author}{Wu, Y.}, \bibinfo{author}{Chen, Z.},
  \bibinfo{author}{Thorat, N.}, \bibinfo{author}{Vi{\'e}gas, F.},
  \bibinfo{author}{Wattenberg, M.}, \bibinfo{author}{Corrado, G.},
  \bibinfo{author}{Hughes, M.}, \bibinfo{author}{Dean, J.},
  \bibinfo{year}{2017}.
\newblock \bibinfo{title}{{G}oogle{'}s multilingual neural machine translation
  system: Enabling zero-shot translation}.
\newblock \bibinfo{journal}{Transactions of the Association for Computational
  Linguistics} \bibinfo{volume}{5}, \bibinfo{pages}{339--351}.
\newblock \DOIprefix\doi{10.1162/tacl_a_00065}.
\bibitem[{Karnati et~al.(2023)Karnati, Seal, Bhattacharjee, Yazidi and
  Krejcar}]{Karnati2023Understanding}
\bibinfo{author}{Karnati, M.}, \bibinfo{author}{Seal, A.},
  \bibinfo{author}{Bhattacharjee, D.}, \bibinfo{author}{Yazidi, A.},
  \bibinfo{author}{Krejcar, O.}, \bibinfo{year}{2023}.
\newblock \bibinfo{title}{Understanding deep learning techniques for
  recognition of human emotions using facial expressions: A comprehensive
  survey}.
\newblock \bibinfo{journal}{IEEE Transactions on Instrumentation and
  Measurement} \bibinfo{volume}{72}, \bibinfo{pages}{1--31}.
\newblock \DOIprefix\doi{10.1109/TIM.2023.3243661}.
\bibitem[{Lewis et~al.(2020)Lewis, Liu, Goyal, Ghazvininejad, Mohamed, Levy,
  Stoyanov and Zettlemoyer}]{lewis2020bart}
\bibinfo{author}{Lewis, M.}, \bibinfo{author}{Liu, Y.}, \bibinfo{author}{Goyal,
  N.}, \bibinfo{author}{Ghazvininejad, M.}, \bibinfo{author}{Mohamed, A.},
  \bibinfo{author}{Levy, O.}, \bibinfo{author}{Stoyanov, V.},
  \bibinfo{author}{Zettlemoyer, L.}, \bibinfo{year}{2020}.
\newblock \bibinfo{title}{{BART}: Denoising sequence-to-sequence pre-training
  for natural language generation, translation, and comprehension}, in:
  \bibinfo{booktitle}{Proceedings of the 58th Annual Meeting of the Association
  for Computational Linguistics}, \bibinfo{publisher}{Association for
  Computational Linguistics}, \bibinfo{address}{Online}. pp.
  \bibinfo{pages}{7871--7880}.
\newblock \DOIprefix\doi{10.18653/v1/2020.acl-main.703}.
\bibitem[{Li et~al.(2021a)Li, Mueller, Qian, Perez, Abualshour, Thabet and
  Ghanem}]{li2021deepgcns}
\bibinfo{author}{Li, G.}, \bibinfo{author}{Mueller, M.}, \bibinfo{author}{Qian,
  G.}, \bibinfo{author}{Perez, I.C.D.}, \bibinfo{author}{Abualshour, A.},
  \bibinfo{author}{Thabet, A.K.}, \bibinfo{author}{Ghanem, B.},
  \bibinfo{year}{2021}a.
\newblock \bibinfo{title}{{DeepGCNs}: Making {GCNs} go as deep as {CNNs}}.
\newblock \bibinfo{journal}{IEEE Transactions on Pattern Analysis and Machine
  Intelligence} ,
  \bibinfo{pages}{1--1}\DOIprefix\doi{10.1109/tpami.2021.3074057}.
\bibitem[{Li et~al.(2020)Li, Ji, Li, Zhang and Liu}]{li2020hitrans}
\bibinfo{author}{Li, J.}, \bibinfo{author}{Ji, D.}, \bibinfo{author}{Li, F.},
  \bibinfo{author}{Zhang, M.}, \bibinfo{author}{Liu, Y.}, \bibinfo{year}{2020}.
\newblock \bibinfo{title}{{H}i{T}rans: A transformer-based context- and
  speaker-sensitive model for emotion detection in conversations}, in:
  \bibinfo{booktitle}{Proceedings of the 28th International Conference on
  Computational Linguistics}, \bibinfo{publisher}{International Committee on
  Computational Linguistics}. pp. \bibinfo{pages}{4190--4200}.
\bibitem[{Li et~al.(2021b)Li, Lin, Fu and Wang}]{li2021pastpresent}
\bibinfo{author}{Li, J.}, \bibinfo{author}{Lin, Z.}, \bibinfo{author}{Fu, P.},
  \bibinfo{author}{Wang, W.}, \bibinfo{year}{2021}b.
\newblock \bibinfo{title}{Past, present, and future: Conversational emotion
  recognition through structural modeling of psychological knowledge}, in:
  \bibinfo{booktitle}{Findings of the Association for Computational
  Linguistics: EMNLP 2021}, \bibinfo{publisher}{Association for Computational
  Linguistics}, \bibinfo{address}{Punta Cana, Dominican Republic}. pp.
  \bibinfo{pages}{1204--1214}.
\newblock \DOIprefix\doi{10.18653/v1/2021.findings-emnlp.104}.
\bibitem[{Li et~al.(2018)Li, Han and ming Wu}]{li2018deeper}
\bibinfo{author}{Li, Q.}, \bibinfo{author}{Han, Z.}, \bibinfo{author}{ming Wu,
  X.}, \bibinfo{year}{2018}.
\newblock \bibinfo{title}{Deeper insights into graph convolutional networks for
  semi-supervised learning}.
\newblock \bibinfo{journal}{Proceedings of the {AAAI} Conference on Artificial
  Intelligence} \bibinfo{volume}{32}.
\newblock \DOIprefix\doi{10.1609/aaai.v32i1.11604}.
\bibitem[{Li et~al.(2022a)Li, Yan and Qiu}]{li2022contrast}
\bibinfo{author}{Li, S.}, \bibinfo{author}{Yan, H.}, \bibinfo{author}{Qiu, X.},
  \bibinfo{year}{2022}a.
\newblock \bibinfo{title}{Contrast and generation make bart a good dialogue
  emotion recognizer}, in: \bibinfo{booktitle}{Proceedings of the AAAI
  Conference on Artificial Intelligence}, pp. \bibinfo{pages}{11002--11010}.
\bibitem[{Li et~al.(2022b)Li, Shao, Ji and Cambria}]{li2022bieru}
\bibinfo{author}{Li, W.}, \bibinfo{author}{Shao, W.}, \bibinfo{author}{Ji, S.},
  \bibinfo{author}{Cambria, E.}, \bibinfo{year}{2022}b.
\newblock \bibinfo{title}{Bieru: Bidirectional emotional recurrent unit for
  conversational sentiment analysis}.
\newblock \bibinfo{journal}{Neurocomputing} \bibinfo{volume}{467},
  \bibinfo{pages}{73--82}.
\newblock \DOIprefix\doi{10.1016/j.neucom.2021.09.057}.
\bibitem[{Li et~al.(2017)Li, Su, Shen, Li, Cao and Niu}]{li2017dailydialog}
\bibinfo{author}{Li, Y.}, \bibinfo{author}{Su, H.}, \bibinfo{author}{Shen, X.},
  \bibinfo{author}{Li, W.}, \bibinfo{author}{Cao, Z.}, \bibinfo{author}{Niu,
  S.}, \bibinfo{year}{2017}.
\newblock \bibinfo{title}{{D}aily{D}ialog: A manually labelled multi-turn
  dialogue dataset}, in: \bibinfo{booktitle}{Proceedings of the Eighth
  International Joint Conference on Natural Language Processing (Volume 1: Long
  Papers)}, \bibinfo{publisher}{Asian Federation of Natural Language
  Processing}, \bibinfo{address}{Taipei, Taiwan}. pp.
  \bibinfo{pages}{986--995}.
\bibitem[{Liang et~al.(2022)Liang, Xu, Lin, Yang and Wang}]{liang2022page}
\bibinfo{author}{Liang, C.}, \bibinfo{author}{Xu, J.}, \bibinfo{author}{Lin,
  Y.}, \bibinfo{author}{Yang, C.}, \bibinfo{author}{Wang, Y.},
  \bibinfo{year}{2022}.
\newblock \bibinfo{title}{{S}+{PAGE}: A speaker and position-aware graph neural
  network model for emotion recognition in conversation}, in:
  \bibinfo{booktitle}{Proceedings of the 2nd Conference of the Asia-Pacific
  Chapter of the Association for Computational Linguistics and the 12th
  International Joint Conference on Natural Language Processing (Volume 1: Long
  Papers)}, \bibinfo{publisher}{Association for Computational Linguistics},
  \bibinfo{address}{Online only}. pp. \bibinfo{pages}{148--157}.
\bibitem[{Liu et~al.(2020)Liu, Ott, Goyal, Du, Joshi, Chen, Levy, Lewis,
  Zettlemoyer and Stoyanov}]{liu2020roberta}
\bibinfo{author}{Liu, Y.}, \bibinfo{author}{Ott, M.}, \bibinfo{author}{Goyal,
  N.}, \bibinfo{author}{Du, J.}, \bibinfo{author}{Joshi, M.},
  \bibinfo{author}{Chen, D.}, \bibinfo{author}{Levy, O.},
  \bibinfo{author}{Lewis, M.}, \bibinfo{author}{Zettlemoyer, L.},
  \bibinfo{author}{Stoyanov, V.}, \bibinfo{year}{2020}.
\newblock \bibinfo{title}{Ro{\{}bert{\}}a: A robustly optimized {\{}bert{\}}
  pretraining approach}.
\bibitem[{Loshchilov and Hutter(2019)}]{loshchilov2018decoupled}
\bibinfo{author}{Loshchilov, I.}, \bibinfo{author}{Hutter, F.},
  \bibinfo{year}{2019}.
\newblock \bibinfo{title}{Decoupled weight decay regularization}, in:
  \bibinfo{booktitle}{International Conference on Learning Representations},
  pp. \bibinfo{pages}{1--8}.
\bibitem[{Majumder et~al.(2019)Majumder, Poria, Hazarika, Mihalcea, Gelbukh and
  Cambria}]{majumder2019dialoguernn}
\bibinfo{author}{Majumder, N.}, \bibinfo{author}{Poria, S.},
  \bibinfo{author}{Hazarika, D.}, \bibinfo{author}{Mihalcea, R.},
  \bibinfo{author}{Gelbukh, A.}, \bibinfo{author}{Cambria, E.},
  \bibinfo{year}{2019}.
\newblock \bibinfo{title}{{DialogueRNN}: An attentive {RNN} for emotion
  detection in conversations}, in: \bibinfo{booktitle}{Proceedings of the AAAI
  Conference on Artificial Intelligence}, \bibinfo{publisher}{Association for
  the Advancement of Artificial Intelligence ({AAAI})}. pp.
  \bibinfo{pages}{6818--6825}.
\newblock \DOIprefix\doi{10.1609/aaai.v33i01.33016818}.
\bibitem[{Nie et~al.(2022)Nie, Chang, Ren, Su and Liu}]{nie2022igcn}
\bibinfo{author}{Nie, W.}, \bibinfo{author}{Chang, R.}, \bibinfo{author}{Ren,
  M.}, \bibinfo{author}{Su, Y.}, \bibinfo{author}{Liu, A.},
  \bibinfo{year}{2022}.
\newblock \bibinfo{title}{I-gcn: Incremental graph convolution network for
  conversation emotion detection}.
\newblock \bibinfo{journal}{IEEE Transactions on Multimedia}
  \bibinfo{volume}{24}, \bibinfo{pages}{4471--4481}.
\newblock \DOIprefix\doi{10.1109/TMM.2021.3118881}.
\bibitem[{Peng et~al.(2020)Peng, Zhou, Liu and Ai}]{peng2020humanmachine}
\bibinfo{author}{Peng, D.}, \bibinfo{author}{Zhou, M.}, \bibinfo{author}{Liu,
  C.}, \bibinfo{author}{Ai, J.}, \bibinfo{year}{2020}.
\newblock \bibinfo{title}{Human–machine dialogue modelling with the fusion of
  word- and sentence-level emotions}.
\newblock \bibinfo{journal}{Knowledge-Based Systems} \bibinfo{volume}{192},
  \bibinfo{pages}{1--9}.
\newblock \DOIprefix\doi{10.1016/j.knosys.2019.105319}.
\bibitem[{Poria et~al.(2019)Poria, Hazarika, Majumder, Naik, Cambria and
  Mihalcea}]{poria2018meld}
\bibinfo{author}{Poria, S.}, \bibinfo{author}{Hazarika, D.},
  \bibinfo{author}{Majumder, N.}, \bibinfo{author}{Naik, G.},
  \bibinfo{author}{Cambria, E.}, \bibinfo{author}{Mihalcea, R.},
  \bibinfo{year}{2019}.
\newblock \bibinfo{title}{{MELD}: A multimodal multi-party dataset for emotion
  recognition in conversations}, in: \bibinfo{booktitle}{Proceedings of the
  57th Annual Meeting of the Association for Computational Linguistics},
  \bibinfo{publisher}{Association for Computational Linguistics},
  \bibinfo{address}{Florence, Italy}. pp. \bibinfo{pages}{527--536}.
\newblock \DOIprefix\doi{10.18653/v1/p19-1050}.
\bibitem[{Ren et~al.(2022)Ren, Huang, Li, Song and Nie}]{ren2022lrgcn}
\bibinfo{author}{Ren, M.}, \bibinfo{author}{Huang, X.}, \bibinfo{author}{Li,
  W.}, \bibinfo{author}{Song, D.}, \bibinfo{author}{Nie, W.},
  \bibinfo{year}{2022}.
\newblock \bibinfo{title}{Lr-gcn: Latent relation-aware graph convolutional
  network for conversational emotion recognition}.
\newblock \bibinfo{journal}{IEEE Transactions on Multimedia}
  \bibinfo{volume}{24}, \bibinfo{pages}{4422--4432}.
\newblock \DOIprefix\doi{10.1109/TMM.2021.3117062}.
\bibitem[{Seal et~al.(2020)Seal, Reddy, Chaithanya, Meghana, Jahnavi, Krejcar
  and Hudak}]{seal2020eeg}
\bibinfo{author}{Seal, A.}, \bibinfo{author}{Reddy, P.P.N.},
  \bibinfo{author}{Chaithanya, P.}, \bibinfo{author}{Meghana, A.},
  \bibinfo{author}{Jahnavi, K.}, \bibinfo{author}{Krejcar, O.},
  \bibinfo{author}{Hudak, R.}, \bibinfo{year}{2020}.
\newblock \bibinfo{title}{An eeg database and its initial benchmark emotion
  classification performance}.
\newblock \bibinfo{journal}{Computational and Mathematical Methods in Medicine}
  \bibinfo{volume}{2020}.
\bibitem[{Shen et~al.(2021)Shen, Chen, Quan and Xie}]{shen2021dialogxl}
\bibinfo{author}{Shen, W.}, \bibinfo{author}{Chen, J.}, \bibinfo{author}{Quan,
  X.}, \bibinfo{author}{Xie, Z.}, \bibinfo{year}{2021}.
\newblock \bibinfo{title}{Dialogxl: All-in-one xlnet for multi-party
  conversation emotion recognition}, in: \bibinfo{booktitle}{Proceedings of the
  AAAI Conference on Artificial Intelligence}, pp.
  \bibinfo{pages}{13789--13797}.
\newblock \DOIprefix\doi{10.1609/aaai.v35i15.17625}.
\bibitem[{Song et~al.(2023)Song, Giunchiglia, Shi, Shen and Xu}]{song2023sunet}
\bibinfo{author}{Song, R.}, \bibinfo{author}{Giunchiglia, F.},
  \bibinfo{author}{Shi, L.}, \bibinfo{author}{Shen, Q.}, \bibinfo{author}{Xu,
  H.}, \bibinfo{year}{2023}.
\newblock \bibinfo{title}{{SUNET}: Speaker-utterance interaction graph neural
  network for emotion recognition in conversations}.
\newblock \bibinfo{journal}{Engineering Applications of Artificial
  Intelligence} \bibinfo{volume}{123}, \bibinfo{pages}{106315}.
\newblock \DOIprefix\doi{10.1016/j.engappai.2023.106315}.
\bibitem[{Song et~al.(2022)Song, Zang, Zhang, Hu and
  Huang}]{song2022emotionflow}
\bibinfo{author}{Song, X.}, \bibinfo{author}{Zang, L.}, \bibinfo{author}{Zhang,
  R.}, \bibinfo{author}{Hu, S.}, \bibinfo{author}{Huang, L.},
  \bibinfo{year}{2022}.
\newblock \bibinfo{title}{Emotionflow: Capture the dialogue level emotion
  transitions}, in: \bibinfo{booktitle}{ICASSP 2022 - 2022 IEEE International
  Conference on Acoustics, Speech and Signal Processing (ICASSP)}, pp.
  \bibinfo{pages}{8542--8546}.
\newblock \DOIprefix\doi{10.1109/ICASSP43922.2022.9746464}.
\bibitem[{Ullah et~al.(2022)Ullah, Qi, Hasan and Asim}]{ullah2022improved}
\bibinfo{author}{Ullah, Z.}, \bibinfo{author}{Qi, L.}, \bibinfo{author}{Hasan,
  A.}, \bibinfo{author}{Asim, M.}, \bibinfo{year}{2022}.
\newblock \bibinfo{title}{Improved deep cnn-based two stream super resolution
  and hybrid deep model-based facial emotion recognition}.
\newblock \bibinfo{journal}{Engineering Applications of Artificial
  Intelligence} \bibinfo{volume}{116}, \bibinfo{pages}{105486}.
\newblock \DOIprefix\doi{10.1016/j.engappai.2022.105486}.
\bibitem[{Vaswani et~al.(2017)Vaswani, Shazeer, Parmar, Uszkoreit, Jones,
  Gomez, Kaiser and Polosukhin}]{vaswani2017attention}
\bibinfo{author}{Vaswani, A.}, \bibinfo{author}{Shazeer, N.},
  \bibinfo{author}{Parmar, N.}, \bibinfo{author}{Uszkoreit, J.},
  \bibinfo{author}{Jones, L.}, \bibinfo{author}{Gomez, A.N.},
  \bibinfo{author}{Kaiser, L.}, \bibinfo{author}{Polosukhin, I.},
  \bibinfo{year}{2017}.
\newblock \bibinfo{title}{Attention is all you need}, in:
  \bibinfo{booktitle}{Advances in Neural Information Processing Systems}, pp.
  \bibinfo{pages}{1--11}.
\bibitem[{Wenxiang~Jiao and King(2020)}]{jiao2020real}
\bibinfo{author}{Wenxiang~Jiao, M.R.L.}, \bibinfo{author}{King, I.},
  \bibinfo{year}{2020}.
\newblock \bibinfo{title}{Real-time emotion recognition via attention gated
  hierarchical memory network}, in: \bibinfo{booktitle}{The Thirty-Fourth
  {AAAI} Conference on Artificial Intelligence, {AAAI} 2020}, pp.
  \bibinfo{pages}{8002--8009}.
\bibitem[{Xiao et~al.(2023)Xiao, Xing, Zhao, Qu, Luo, Dai, Li and
  Zhu}]{xiao2023deep}
\bibinfo{author}{Xiao, Z.}, \bibinfo{author}{Xing, H.}, \bibinfo{author}{Zhao,
  B.}, \bibinfo{author}{Qu, R.}, \bibinfo{author}{Luo, S.},
  \bibinfo{author}{Dai, P.}, \bibinfo{author}{Li, K.}, \bibinfo{author}{Zhu,
  Z.}, \bibinfo{year}{2023}.
\newblock \bibinfo{title}{Deep contrastive representation learning with
  self-distillation}.
\newblock \bibinfo{journal}{IEEE Transactions on Emerging Topics in
  Computational Intelligence} ,
  \bibinfo{pages}{1--13}\DOIprefix\doi{10.1109/TETCI.2023.3304948}.
\bibitem[{Xie et~al.(2021)Xie, Yang, Sun, Liu and
  Ji}]{xie2021knowledgeinteractive}
\bibinfo{author}{Xie, Y.}, \bibinfo{author}{Yang, K.}, \bibinfo{author}{Sun,
  C.}, \bibinfo{author}{Liu, B.}, \bibinfo{author}{Ji, Z.},
  \bibinfo{year}{2021}.
\newblock \bibinfo{title}{Knowledge-interactive network with sentiment polarity
  intensity-aware multi-task learning for emotion recognition in
  conversations}, in: \bibinfo{booktitle}{Findings of the Association for
  Computational Linguistics: EMNLP 2021}, \bibinfo{publisher}{Association for
  Computational Linguistics}, \bibinfo{address}{Punta Cana, Dominican
  Republic}. pp. \bibinfo{pages}{2879--2889}.
\newblock \DOIprefix\doi{10.18653/v1/2021.findings-emnlp.245}.
\bibitem[{Xing et~al.(2022)Xing, Xiao, Zhan, Luo, Dai and
  Li}]{xing2022SelfMatch}
\bibinfo{author}{Xing, H.}, \bibinfo{author}{Xiao, Z.}, \bibinfo{author}{Zhan,
  D.}, \bibinfo{author}{Luo, S.}, \bibinfo{author}{Dai, P.},
  \bibinfo{author}{Li, K.}, \bibinfo{year}{2022}.
\newblock \bibinfo{title}{{SelfMatch}: Robust semisupervised time-series
  classification with self-distillation}.
\newblock \bibinfo{journal}{International Journal of Intelligent Systems}
  \bibinfo{volume}{37}, \bibinfo{pages}{8583--8610}.
\newblock \DOIprefix\doi{10.1002/int.22957}.
\bibitem[{Xu et~al.(2022)Xu, Yuan, Zhao, Xu, Zou and Gao}]{xu2022gar}
\bibinfo{author}{Xu, H.}, \bibinfo{author}{Yuan, Z.Q.}, \bibinfo{author}{Zhao,
  K.}, \bibinfo{author}{Xu, Y.F.}, \bibinfo{author}{Zou, J.Y.},
  \bibinfo{author}{Gao, K.}, \bibinfo{year}{2022}.
\newblock \bibinfo{title}{{GAR-Net}: A graph attention reasoning network for
  conversation understanding}.
\newblock \bibinfo{journal}{Knowledge-Based Systems} \bibinfo{volume}{240},
  \bibinfo{pages}{108055}.
\bibitem[{Yang et~al.(2019)Yang, Dai, Yang, Carbonell, Salakhutdinov and
  Le}]{yang2019xlnet}
\bibinfo{author}{Yang, Z.}, \bibinfo{author}{Dai, Z.}, \bibinfo{author}{Yang,
  Y.}, \bibinfo{author}{Carbonell, J.}, \bibinfo{author}{Salakhutdinov, R.},
  \bibinfo{author}{Le, Q.V.}, \bibinfo{year}{2019}.
\newblock \bibinfo{title}{Xlnet: Generalized autoregressive pretraining for
  language understanding}, in: \bibinfo{booktitle}{Proceedings of the 33rd
  International Conference on Neural Information Processing Systems},
  \bibinfo{publisher}{Curran Associates Inc.}, \bibinfo{address}{Red Hook, NY,
  USA}. pp. \bibinfo{pages}{1--11}.
\bibitem[{Zahiri and Choi(2018)}]{zahiri2018emotion}
\bibinfo{author}{Zahiri, S.M.}, \bibinfo{author}{Choi, J.D.},
  \bibinfo{year}{2018}.
\newblock \bibinfo{title}{Emotion detection on tv show transcripts with
  sequence-based convolutional neural networks}, in: \bibinfo{booktitle}{The
  Workshops of the The Thirty-Second AAAI Conference on Artificial
  Intelligence}, pp. \bibinfo{pages}{44--52}.
\bibitem[{Zhang et~al.(2020)Zhang, Lu, Sak, Tripathi, McDermott, Koo and
  Kumar}]{zhang2020transformer}
\bibinfo{author}{Zhang, Q.}, \bibinfo{author}{Lu, H.}, \bibinfo{author}{Sak,
  H.}, \bibinfo{author}{Tripathi, A.}, \bibinfo{author}{McDermott, E.},
  \bibinfo{author}{Koo, S.}, \bibinfo{author}{Kumar, S.}, \bibinfo{year}{2020}.
\newblock \bibinfo{title}{Transformer transducer: A streamable speech
  recognition model with transformer encoders and {RNN}-t loss}, in:
  \bibinfo{booktitle}{Proceedings of 2020 {IEEE} International Conference on
  Acoustics, Speech and Signal Processing ({ICASSP})},
  \bibinfo{organization}{IEEE}. \bibinfo{publisher}{{IEEE}}. pp.
  \bibinfo{pages}{7829--7833}.
\newblock \DOIprefix\doi{10.1109/icassp40776.2020.9053896}.
\bibitem[{Zhao et~al.(2022)Zhao, Zhao and Lu}]{zhao2022cauain}
\bibinfo{author}{Zhao, W.}, \bibinfo{author}{Zhao, Y.}, \bibinfo{author}{Lu,
  X.}, \bibinfo{year}{2022}.
\newblock \bibinfo{title}{Cauain: Causal aware interaction network for emotion
  recognition in conversations}, in: \bibinfo{editor}{Raedt, L.D.} (Ed.),
  \bibinfo{booktitle}{Proceedings of the Thirty-First International Joint
  Conference on Artificial Intelligence, {IJCAI-22}},
  \bibinfo{publisher}{International Joint Conferences on Artificial
  Intelligence Organization}. pp. \bibinfo{pages}{4524--4530}.
\newblock \DOIprefix\doi{10.24963/ijcai.2022/628}. \bibinfo{note}{main Track}.
\bibitem[{Zhu et~al.(2021)Zhu, Pergola, Gui, Zhou and He}]{zhu2021topic}
\bibinfo{author}{Zhu, L.}, \bibinfo{author}{Pergola, G.}, \bibinfo{author}{Gui,
  L.}, \bibinfo{author}{Zhou, D.}, \bibinfo{author}{He, Y.},
  \bibinfo{year}{2021}.
\newblock \bibinfo{title}{Topic-driven and knowledge-aware transformer for
  dialogue emotion detection}, in: \bibinfo{booktitle}{Proceedings of the 59th
  Annual Meeting of the Association for Computational Linguistics and the 11th
  International Joint Conference on Natural Language Processing (Volume 1: Long
  Papers)}, \bibinfo{publisher}{Association for Computational Linguistics},
  \bibinfo{address}{Online}. pp. \bibinfo{pages}{1571--1582}.
\newblock \DOIprefix\doi{10.18653/v1/2021.acl-long.125}.

\end{thebibliography}

\end{document}